%% file: main.tex
\documentclass[letterpaper]{article} 
\usepackage{aaai2027}  
\usepackage[hyphens]{url}  
\usepackage{graphicx} 
\usepackage{natbib}  
\usepackage{caption} 
\usepackage{amsmath}
\usepackage{amssymb}
\usepackage{booktabs}

\title{How Far Do Foundation Models Transfer to Infant Signals?\\A Cross-Dataset Transfer Audit with a Unified Need Ontology}
\author{Wu Hangyu}
\affiliations{
Shenzhen Coddie Technology Co., Ltd.
}

\begin{document}
\maketitle

\begin{abstract}
Public infant cry corpora are small, label-incompatible, and almost always evaluated one corpus at a time. We ask what this practice hides---and what fixes it. Across four cry corpora screened by a multi-level leakage audit (byte-level and embedding-level deduplication), we probe four frozen encoders and a handcrafted baseline under a unified five-class need ontology and shared task formulations. The audit exposes what single-corpus evaluation conceals: within-domain macro-F1 swings by 0.52--0.77 for the same encoder, cross-corpus transfer is negative on average (negative-transfer ratio 0.18--0.34, significant in 17 of 30 directed cells, BH-FDR), and 349 content-identical clip groups carry conflicting labels across corpora. The same audit, however, reveals a consistent way forward. Transfer into the noisiest corpus is reliably positive---at matched training size and after near-duplicate removal---offering a practical recipe for small, noisy corpora. Frozen probes saturate at modest label budgets, while stabilized fine-tuning wins with full labels; domain-adaptive pretraining helps mainly at the full-data end. Above all, ontology-mapped joint training wins in all four encoder-by-target settings, whereas naively merging unmapped labels costs up to 37 F1 points. We release the ontology, mapping code, and audit pipeline, turning incompatible cry corpora into a usable joint-training resource.
\end{abstract}

\input{sections/intro}
\input{sections/related}
\input{sections/method}
\input{sections/experiments}
\input{sections/conclusion}

\bibliography{references}

\clearpage
\appendix
\setcounter{section}{0}
\setcounter{table}{0}
\setcounter{figure}{0}
\renewcommand{\thetable}{A\arabic{table}}
\renewcommand{\thefigure}{A\arabic{figure}}
\renewcommand{\thesection}{A\arabic{section}}

\input{sections/appendix}

\end{document}

%% file: sections/intro.tex
\section{Introduction}
\label{sec:intro}

Caregiving technology increasingly promises to tell parents \emph{why} an infant is crying — hungry, tired, in discomfort, or in need of soothing. Behind that promise sits a small research field built on a handful of public cry corpora, each collected under different conditions, from different populations, with different label conventions. As audio foundation models become the default feature extractors for such tasks, a practical question arises for anyone who trains on this data: how far do these models transfer to infant signals?

Infant cry classification has progressed from MFCC-and-SVM pipelines to deep architectures \citep{maghfira2020cnnrnn,ozseven2023cry,zayed2023cry}, and reviews agree on the obstacle: not model capacity, but data \citep{ji2021review,hashemi2025survey}. Public corpora contain hundreds of clips, not millions; labels are caregiver inferences from context rather than physiological ground truth \citep{gustafson1990cries,wood2001infant}; and each corpus defines its own label set, so the data cannot even be pooled without semantic decisions. Meanwhile, evaluation remains almost entirely within-corpus, under protocols that differ from paper to paper.

What the field lacks is not another classifier but a measurement of the assumptions its pipelines rest on. Does a frozen foundation encoder represent cry equally well across corpora? Does training on one corpus help on another, or hurt? Do general audio MLLMs already solve the problem zero-shot? Does domain-adaptive pretraining pay off at realistic label budgets? And what happens if the corpora are naively merged despite their incompatible labels?

The need for such an audit is easy to demonstrate. Recomputing content hashes across the raw, pre-deduplication versions of the four corpora we study, we find $349$ groups of acoustically identical clips that carry different labels in different corpora — $292$ of them ($83.7\%$) disputes about whether a cry means \emph{hungry}. These are \emph{cross-distribution metadata/label conflicts}: some plausibly reflect independent re-annotation of the same recording, others dataset repackaging, directory naming, or secondary rule mapping, and for a substantial fraction the provenance cannot be verified from the released files (Section~\ref{sec:ontology}); our released manifest keeps only one copy of each group, so these conflicts are a label-convention audit, not leakage. And when the same frozen encoder is probed within each corpus, macro-F1 swings by $0.57$--$0.80$ depending on which corpus supplies the labels. Cross-corpus work that ignores these facts risks reporting leakage as transfer and convention as semantics.

This paper performs the audit — a \emph{multi-corpus foundation-encoder transfer audit with ontology and leakage controls}. We define a five-class need ontology over the four corpora, with explicit mapping rules and a multi-level leakage audit (byte-level deduplication, embedding-level near-duplicate screening, and a within-corpus train--test near-duplicate audit); we measure a train$\times$test transfer matrix for four audio foundation encoders and a handcrafted baseline under a unified held-out-test protocol, summarized with a negative-transfer ratio, cluster-bootstrap tests, and FDR control; we sweep label budgets under three adaptation arms; and we measure what the ontology buys in joint training — and what skipping it costs.

Our findings are diagnostic rather than triumphalist: \textbf{(i)} foundation encoders show a corpus-level domain gap of up to $0.80$ macro-F1, and on the hardest corpus a handcrafted XGBoost baseline beats two of the four deep encoders and matches a third, while a zero-shot general audio MLLM --- probed as a sixth, training-free ``encoder'' (Appendix~\ref{app:mllm}) --- trails even the weakest in-domain probe; \textbf{(ii)} negative transfer is the norm (NTR $0.19$--$0.35$, up to $0.46$ cry-only) and asymmetric, with one instructive exception — transfer \emph{into} the noisiest, lowest-diagonal corpus is consistently positive in effect size, including after near-duplicate removal, though not individually significant; \textbf{(iii)} frozen probes saturate at modest label budgets, DAPT significantly beats stabilized fine-tuning at $5$--$10$-shot (the 1-shot advantage is not robust to optimization-seed variance), the 50-shot crossover to stabilized fine-tuning is a non-significant trend (paired 95\% CI includes zero) that materializes only at the full-data budget, the 1-shot arm differences lie within shot-sampling noise, and naive full fine-tuning collapses; \textbf{(iv)} in the tested binary, shared-label settings, ontology-mapped joint training is best in all four encoder$\times$target combinations, while merging unmapped labels costs up to $37$ F1 points; and \textbf{(v)} a pre-specified quality--transfer correlation is refuted as stated — clip-level signal quality acts only as a proxy for domain identity. We release the ontology, mapping code, deduplication manifests, and audit pipeline.

Section~\ref{sec:related} situates the audit; Section~\ref{sec:method} formalizes it; Section~\ref{sec:experiments} reports the results; Section~\ref{sec:conclusion} concludes with limitations.

%% file: sections/related.tex
\section{Related Work}
\label{sec:related}

\textbf{Infant cry analysis.}
Cry classification has moved from handcrafted features with SVMs to CNN--RNN and other deep classifiers \citep{maghfira2020cnnrnn,ozseven2023cry,vincent2021neonatal,zayed2023cry,cohen2019baby}, with cry \emph{detection} in domestic audio as a parallel line \citep{lavner2016cry}. Reviews consistently name the same bottleneck: small, clinically skewed corpora, labels inferred from context, and poor cross-device generalization \citep{ji2021review,hashemi2025survey}; corpora are rarely combined \citep{sharma2015database,laguna2023multimodal,chittora2017collection}. We do not propose a new classifier; we measure how far existing foundation encoders travel across the corpora the field already has.

\textbf{Cross-corpus transfer and negative transfer.}
Cross-corpus degradation is well documented in speech emotion recognition \citep{schuller2010crosscorpus}, and transfer matrices are known to be asymmetric \citep{milner2019crosscorpus}; negative transfer is a general phenomenon of transfer learning \citep{pan2010survey,zamir2018taskonomy}. \emph{Known from cross-corpus SER}: transfer degrades across corpora, matrices are asymmetric, and corpus identity confounds evaluation \citep{braunschweiler2021corpus}. Cross-corpus infant-cry classification, domain adaptation, and pretrained-audio transfer onto infant vocalization all have prior art; we do not claim to be first to study cross-domain infant cry or to use pretrained audio representations. \emph{What this audit adds} is the combination --- a multi-corpus foundation-encoder transfer audit with ontology and leakage controls: (i) a full train$\times$test transfer matrix on infant cry under a unified label space \emph{and} a unified held-out-test protocol with cluster-bootstrap significance and FDR control; (ii) a signed exception analysis — reverse transfer \emph{into} the noisiest, lowest-diagonal corpus, consistent in effect size and verified against near-duplicate leakage; (iii) a quantified cost of label-ontology mismatch (naive merge vs.\ mapped joint); and (iv) a multi-level leakage audit --- cross-domain and within-corpus --- with removal-robustness: methodological pieces that cross-corpus studies typically assume rather than measure.

\textbf{Audio foundation models and frozen evaluation.}
Self-supervised speech encoders \citep{baevski2020wav2vec2,chen2022wavlm} and AudioSet-pretrained models \citep{gong2021ast,wu2023clap,gemmeke2017audioset} are now default feature extractors, typically benchmarked under frozen probes \citep{yang2021superb,pasad2021layerwise}. General AudioSet models still cover non-speech vocalizations poorly unless in-domain vocalization data is added \citep{gong2022vocalsound}; speech SSL degrades under domain shift and recovers with in-domain continued pretraining \citep{hsu2021robust}; infant-cry SSL has been attempted at small scale \citep{gorin2023cryssl}. General audio--language MLLMs \citep{chu2023qwenaudio,xu2025qwenomni,xu2025qwen3omni} add instruction following on top of audio understanding; we evaluate such a model as a zero-shot baseline rather than as a fine-tuning target. We keep all encoders frozen and off-the-shelf, and ask the deployment question — which pretraining source travels best to infant cry.

\textbf{Multi-dataset label-space unification.}
In vision, multi-dataset training forced explicit taxonomy unification \citep{lambert2020mseg,zhou2023unidet,shao2019objects365,wang2019unify}, and the data-engineering literature warns that annotation decisions cascade silently into model failures \citep{gebru2021datasheets,sambasivan2021cascades,northcutt2021labelerrors,fonseca2019noisy,song2022noisylabel}. Our setting differs in kind: cry need labels are caregiver inferences, not perceptual ground truth \citep{gustafson1990cries,wood2001infant,lingle2012cry}, so conflicts are expected; we quantify them ($349$ content-identical groups) and measure the cost of ignoring them.

\textbf{Adaptation under label scarcity.}
Domain-adaptive pretraining helps most when target labels are scarce \citep{gururangan2020dapt}; full fine-tuning can distort pretrained features out-of-distribution \citep{kumar2022finetuning,wortsman2022wise}. We test these predictions on infant cry with a pre-specified label-budget sweep (dated internal analysis plan), and report a case they anticipate but rarely document end-to-end: the naive fine-tuning arm collapses with high variance exactly where it should be strongest.

\textbf{Infant monitoring beyond audio.}
Clinical monitoring has matured around contactless vitals \citep{mcduff2023camera} and multimodal pain assessment \citep{zamzmi2018pain}; consumer products frame sensing as caregiving support \citep{wang2017quantified}. These lines do not share label spaces with cry corpora; we therefore scope this audit to the audio domains.

%% file: sections/method.tex
\section{Method}
\label{sec:method}

\subsection{Formulation}
\label{sec:formulation}

\textbf{Corpora and label spaces.}
Let $\mathcal{D}=\{D_1,\dots,D_K\}$ be $K$ infant-cry corpora. Corpus $D_k$ contains clips $(x,y,g)$ with waveform $x$, raw label $y\in\mathcal{L}_k$, and group identifier $g$ recovered from the available metadata (an infant, recording session, device, or file group, depending on the corpus; per-corpus semantics in Section~\ref{sec:within}). The raw label spaces $\mathcal{L}_k$ are mutually inconsistent: the same acoustic content may be labeled \emph{hungry} in one corpus and \emph{discomfort} in another. We therefore treat each corpus as a separate \emph{domain} and never merge labels without an explicit mapping.

\textbf{Transfer matrix.}
Given a fixed encoder $f$ (frozen) and a fixed probe protocol, the transfer matrix $T\in\mathbb{R}^{K\times K}$ is
\begin{equation}
\label{eq:tmatrix}
T[i,j]=\mathrm{F1}\!\left(\,\mathrm{probe}\!\left(f,\,D_i^{\mathrm{train}}\right)\ \text{evaluated on } D_j\,\right),
\end{equation}
where $\mathrm{F1}$ is macro-F1. \emph{All} entries are evaluated on the held-out test split of the target domain $D_j$ (unified protocol): off-diagonal models are trained on source-domain clips only and never touch target-domain data, so this introduces no leakage while keeping the NTR numerator and denominator directly comparable.

\textbf{Negative-transfer ratio.}
We summarize the matrix by the negative-transfer ratio
\begin{equation}
\label{eq:ntr}
\mathrm{NTR}=1-\frac{\frac{1}{K(K-1)}\sum_{i\neq j}T[i,j]}{\frac{1}{K}\sum_{i}T[i,i]}.
\end{equation}
$\mathrm{NTR}>0$ means training on another corpus is on average worse than training in-domain. Because each cell is produced by the full probe pipeline, NTR is an \emph{end-to-end protocol transfer deficit}, not a pure measure of representation quality: it mixes training-set size, class priors, in-domain difficulty, label composition, and hyperparameter selection. We also report the pairwise asymmetry $A(i,j)=|T[i,j]-T[j,i]|$.

\textbf{Few-shot frontier.}
For a target domain $j$, an adaptation arm $a$, and a label budget of $n$ clips per class, let $P_a(n)$ be the macro-F1 on $D_j^{\mathrm{test}}$, averaged over seeds. Sweeping $n\in\{1,5,10,50,\mathrm{full}\}$ gives the few-shot efficiency frontier $P_a(n)$; the quantities of interest are which arm wins at small and large $n$, and where the ranking crosses. Sampling protocol: we use a \emph{multi-episode confirmatory protocol} --- an independent stratified shot subset is drawn per episode ($20$ episodes for the deterministic arms A/C, $10$ for the GPU fine-tuning arm Bs), each across three optimization seeds ($42/43/44$; arms A/C are exactly deterministic in the optimization seed, so their episode variance is pure shot-sampling variance, while Bs optimization variance is modeled); we report episode means $\pm$ std and paired-episode confidence intervals for arm differences, seed-averaged within each episode as the conservative headline. The single fixed-subset 3-seed sweep is retained (Table~\ref{tab:fewshot}) for continuity with the full-budget runs. Two caveats accompany this protocol: the 1-shot arm ordering is sensitive to how optimization-seed pairs are counted (Section~\ref{sec:fewshot}), and Arm~C's DAPT stage sees a fixed pool of unlabeled target-domain (and other) waveforms that the label-budget axis does not count, so the arms' \emph{total} data budgets are asymmetric by design (fixed target-unlabeled exposure; Section~\ref{sec:fewshot}).

\textbf{Need ontology.}
An ontology is a pair $\mathcal{O}=(\mathcal{C},\{\phi_k\})$ of $M$ need classes $\mathcal{C}$ and per-corpus maps $\phi_k:\mathcal{L}_k\to\mathcal{C}\cup\{\bot\}$, where $\bot$ marks raw labels excluded from analysis. Cross-corpus training is only ever performed in the mapped space. A \emph{label conflict} is a set of clips with identical audio content (sha1-matched) that receives different mapped labels in different corpora; we report these as \emph{cross-distribution metadata/label conflicts} --- they mix independent re-annotation, repackaging, and rule-mapping provenance (Section~\ref{sec:ontology}) --- and they measure how much of the apparent label diversity is convention rather than acoustics.

\textbf{Claims.}
The audit is organized around six falsifiable claims: \textbf{C1} foundation encoders show a large corpus-level domain gap on infant cry; \textbf{C2} cross-corpus transfer is negative on average ($\mathrm{NTR}>0$) and asymmetric; \textbf{C3} domain-adaptive pretraining dominates naive fine-tuning in the low-label regime, and the ranking reverses with abundant labels; \textbf{C4} ontology-mapped joint training beats single-corpus training and naive label merging on a shared-label evaluation; \textbf{C5} signal quality correlates with transferability; \textbf{C6} raw cross-corpus label conflicts are common, so an explicit ontology is a prerequisite, not a convenience.

\subsection{The Need Ontology}
\label{sec:ontology}

\textbf{Classes.}
We define five classes covering the label inventory of all four corpora: \emph{hunger}, \emph{pain-discomfort}, \emph{sleepiness}, \emph{need-soothing}, and \emph{neutral} (non-cry vocalizations and noise), following the developmental literature, which holds that cry acoustics encode graded distress rather than discrete causes, and that need labels are caregiver inferences from context \citep{gustafson1990cries,wood2001infant,lingle2012cry}. Operationally, the prediction target throughout this paper is therefore the \emph{caregiver-annotated behavioral need} — an annotation convention, not a physiological ground truth; our transfer measurements quantify how far these conventions travel.

\textbf{Mapping rules.}
All mappings are deterministic and released as code. They are author-set rules: no independent double annotation, disagreement-rate measurement, or blind review was performed, and no mapping-sensitivity analysis (leave-one-mapping-out, alternative mappings, or a label-name-normalization intermediate baseline) was run; this bounds how strongly C4 results can be attributed to the ontology (Limitations). The main decisions are: \emph{cold\_hot}$\to$\emph{pain-discomfort}, following the original corpus curation; \emph{burping}$\to$\emph{need-soothing}, read as a request for a caregiving action; \emph{unknown}/\emph{dk}$\to\bot$; \emph{laugh}/\emph{noise}$\to$\emph{neutral}. Because \emph{burping} is the least obvious decision, we flag every analysis that depends on it and treat results involving \emph{need-soothing} as restricted subsets.

\begin{figure*}[t]
\centering
\includegraphics[width=0.64\textwidth]{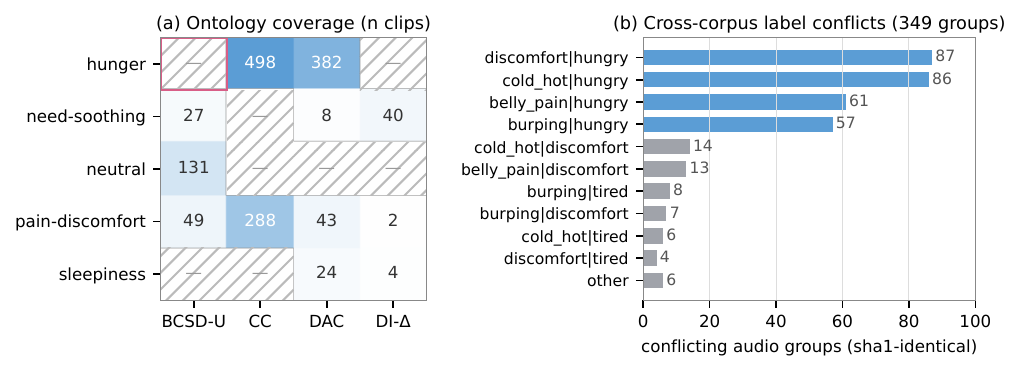}
\caption{\textbf{(a) Ontology coverage.} Sample counts of the five need-ontology classes across the four corpora; hatched ``---'' cells are empty (e.g., \emph{hunger} in BCSD-U; \emph{neutral} exists only in BCSD-U). \textbf{(b) Cross-corpus label conflicts.} The 349 content-identical audio groups with different raw labels across corpus distributions, grouped by label pair; 292 (83.7\%) involve a \emph{hungry} disagreement. Provenance is mixed (re-annotation / repackaging / rule mapping; Section~\ref{sec:ontology}), so these are cross-distribution metadata/label conflicts, not verified independent re-annotations.}
\label{fig:ontology}
\end{figure*}

\textbf{Coverage and conflicts.}
Figure~\ref{fig:ontology}(a) shows the resulting coverage, sparse by construction of the source corpora: \emph{hunger} is absent from BCSD-U after deduplication, \emph{sleepiness} exists only in DAC and DI-$\Delta$, and \emph{neutral} exists only in BCSD-U; empty cells are carried through every downstream analysis rather than silently dropped. Figure~\ref{fig:ontology}(b) shows the conflict audit: computed on the \emph{raw, pre-deduplication} corpus collection, $349$ groups of content-identical audio carry different raw labels across corpus distributions, and $292$ of them ($83.7\%$) are \emph{hungry} vs.\ non-\emph{hungry} disputes (C6) — silently contradictory training signal under naive merging. We deliberately call these \emph{cross-distribution metadata/label conflicts}: some plausibly come from independent re-annotation of shared source recordings, others from repackaging, directory naming, or secondary rule mapping, and the per-source split cannot be verified from the released files (Appendix~\ref{app:ontology}), so the count is an upper bound on independent re-annotation, not a measurement of $349$ caregiver labeling events. Within the released manifest, every conflict group survives only as its DAC copy (Section~\ref{sec:setup}), so the conflicts measure convention, not train/test leakage.

\subsection{Encoders and Adaptation Protocols}
\label{sec:encoders}

\textbf{Frozen encoders.}
We audit four publicly released audio encoders that span the two dominant pretraining paradigms: wav2vec~2.0 base \citep{baevski2020wav2vec2} and WavLM base+ \citep{chen2022wavlm} (self-supervised speech), and CLAP \citep{wu2023clap} and AST \citep{gong2021ast} (AudioSet-pretrained \citep{gemmeke2017audioset}). AST enters as the pre-specified fallback for a fifth planned encoder, voc2vec, whose checkpoint download returned HTTP~401 (gated repository). As non-deep references we compute a 252-dimensional handcrafted set (MFCCs, prosody, spectral statistics) with logistic regression or XGBoost \citep{chen2016xgboost}. All deep encoders are used strictly off-the-shelf; no infant-cry model is trained by us, by design. Exact checkpoints, layers, pooling, and probe hyperparameters are given in the encoder card (Table~\ref{tab:encodercard}, Appendix~\ref{app:encodercard}).

\textbf{Arm A: frozen probe.}
Embeddings are mean-pooled over time and fed to a balanced logistic-regression probe; the regularization constant is selected on the source-domain validation split. This is the pre-specified primary protocol for all transfer measurements, following frozen-evaluation practice in SUPERB \citep{yang2021superb}. An external code audit of this revision cycle found that our original selection loop enumerated the $C$ grid without passing $C$ to the estimator, so every reported run had silently used the default $C{=}1$; we fixed the shared trainer, added unit tests verifying that grid candidates actually change the fitted estimator, and re-ran every probe-dependent result from the cached embeddings. With the corrected selector, non-default values are genuinely chosen in several settings (e.g., $C{=}10$ for mapped joint training and the handcrafted DAC probe; $C{=}0.01$ for many transfer cells); all numbers in this paper are from the corrected runs.

\textbf{Arm B: full fine-tuning.}
All encoder weights are updated on target-domain labels (lr $10^{-4}$, bf16, at most 5 epochs, early stopping on validation); pre-specified in the dated internal analysis plan as the fine-tuning reference.

\textbf{Arm C: domain-adaptive pretraining + probe.}
The encoder is first continued on pooled unlabeled cry audio with the masked-prediction objective (DAPT; \citealp{gururangan2020dapt}) for one epoch, then probed as in Arm A. The unlabeled pool combines the \emph{train split only} of the four corpora ($1{,}037$ clips) with $3{,}500$ clips of CryCeleb \citep{budaghyan2024cryceleb}, a 26k-clip infant-cry corpus released for research use ($4{,}537$ clips in total, $283$ steps). Target validation/test waveforms are excluded from DAPT (\emph{inductive} protocol; Appendix~\ref{app:protocol}).

\textbf{Arm Bs: stabilized fine-tuning (post-hoc diagnostic).}
After Arm B exhibited run collapse (Section~\ref{sec:fewshot}), we added a clearly-marked post-hoc arm that freezes the convolutional feature encoder and fine-tunes the transformer stack at lr $10^{-5}$. All conclusions about the high-label regime are drawn from Bs, and Arm B numbers are reported as an honest failure record rather than as evidence.

\textbf{Zero-shot MLLM probe (post-hoc).}
As a post-hoc control with the same status as Arm Bs, we additionally query a general audio--language MLLM \citep{xu2025qwen3omni} zero-shot on the B1 test splits. Because its data and inference conditions are deliberately asymmetric with the frozen-probe arms (zero-shot vs.\ in-domain supervision; subsampled test sets; API versioning outside our control), all MLLM results and protocol details are reported in Appendix~\ref{app:mllm} only; no claim depends on them.

\subsection{Implementation Details}
\label{sec:impl}

\textbf{Preprocessing and splits.}
All audio is resampled to 16\,kHz mono, cut or padded to 7\,s, and peak-normalized (processing failures: zero). The deduplicated manifest contains $1{,}496$ clips, split group-stratified (by $g$) into train/validation/test $=1037/227/232$; groups never cross splits.

\textbf{Multi-level leakage audit.}
Pairs of files are matched on a dual key (content sha1, filename); $1{,}247$ duplicates are removed before any split ($781$ from BCSD, $466$ from DeepInfant). This is essential: BCSD's \emph{hungry} set is byte-identical to DAC's, and DeepInfant V1 is a repackaging of DAC; without deduplication every cross-corpus number would be inflated by leakage. Byte-level hashing, however, cannot catch re-encoded or trimmed copies, whose hashes differ. We therefore complement it with an embedding-level near-duplicate audit (CLAP cosine nearest neighbors across all domain pairs, second-encoder confirmation), reported with its findings and removal-robustness in Section~\ref{sec:fewshot} and Appendix~\ref{app:leakage}. In the released manifest, cross-domain byte-identical duplicates are zero.

\textbf{Statistics.}
All runs use 3 seeds (scikit-learn \citep{pedregosa2011scikit}); we report mean$\pm$std. For the deterministic arms (frozen features + logistic regression) seed-to-seed std is $\approx0$ by construction; we mark such rows ``deterministic''. Directed transfer cells are tested against the diagonal with a paired \emph{cluster bootstrap} over \texttt{group\_id} ($10{,}000$ resamples; the resampling unit is the recovered-metadata group, not the clip --- per-corpus group semantics and their limits are audited in Section~\ref{sec:within}), with Benjamini--Hochberg FDR control at $q=0.05$ across the $30$ cells.

\textbf{Hardware and budget.}
All experiments run on a single RTX 3090. The DAPT budget is deliberately light (one epoch, 283 steps, 96\,s); Arm C scores are therefore a lower bound, and the DAPT dose--response is left to future work.

\textbf{Ethics and licensing.}
This work targets non-medical caregiving support: the ontology describes behavioral need categories, not clinical states. CryCeleb \citep{budaghyan2024cryceleb} is used as unlabeled DAPT audio under the license terms as stated by its distributors (CC BY-NC-ND 4.0, research use only); we redistribute no CryCeleb audio and release \emph{no derivative model weights} trained on it.

%% file: sections/experiments.tex
\section{Experiments}
\label{sec:experiments}

\subsection{Setup and In-Domain Audit}
\label{sec:setup}

\textbf{Corpora.}
After deduplication, four independent cry corpora remain (Table~\ref{tab:corpora}): the donateacry corpus \citep{donateacry} (DAC; crowdsourced, 8\,kHz), the deduplicated DeepInfant increment \citep{deepinfant} (DI-$\Delta$), the Baby-Crying-Sound-Dataset with DAC-identical files removed \citep{bcsd} (BCSD-U), and CryClass with augmentation groups kept together \citep{cryclass} (CC). DI-$\Delta$ falls to $46$ clips after deduplication and ontology mapping, below our pre-specified $80$-clip threshold; the main transfer matrix is $3\times3$ and DI-$\Delta$ enters only the small-corpus case study (Appendix~\ref{app:didelta}). The $349$ cross-distribution metadata/label-conflict groups of Section~\ref{sec:ontology} are computed on the \emph{raw, pre-deduplication} corpus collection; within the released manifest, cross-domain byte-identical duplicates are zero, and every conflict group survives only as its DAC copy ($341$ sha1 $+8$ filename groups, $346$ clips). Signal quality differs sharply across domains (Section~\ref{sec:quality}).

\begin{table}[t]
\centering
\footnotesize
\caption{Corpus audit after deduplication and ontology mapping (provenance in Table~\ref{tab:appcorpora}). \#Cl.: ontology classes present; Maj.: macro-F1 of a majority-vote classifier (imbalance reference for Table~\ref{tab:b1}); SNR: clip-level median (caveat in Section~\ref{sec:quality}).}
\label{tab:corpora}
\setlength{\tabcolsep}{3pt}
\begin{tabular}{@{}lcccccl@{}}
\toprule
Corpus & Clips & \#Cl. & Maj. & Rate & SNR (dB) & License \\
\midrule
BCSD-U & 207 & 3 & 0.258 & 44.1\,k & 102.9 & unspecified \\
CC & 786 & 2 & 0.388 & 8\,k & 37.2 & unspecified \\
DAC & 457 & 4 & 0.228 & 8\,k & 32.3 & ODbL-1.0 \\
DI-$\Delta$ & 46 & 3 & 0.310 & 44.1\,k & 35.9 & Apache-2.0 \\
\bottomrule
\end{tabular}
\end{table}

\textbf{Pipeline sanity.}
A handcrafted-feature logistic-regression sanity run on DAC reaches macro-F1 $0.192$, and a 64-clip overfit check reaches train accuracy $1.0$ --- the pipeline can memorize small data and has no label leak.

\begin{figure}[t]
\centering
\includegraphics[width=0.62\columnwidth]{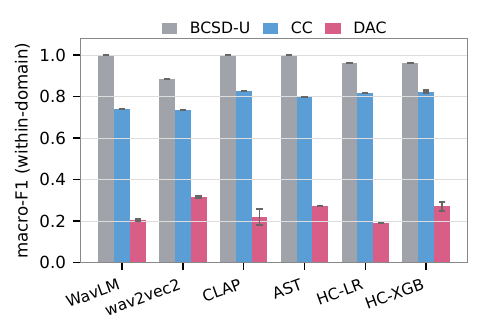}
\caption{\textbf{Within-domain performance varies by $>0.5$ macro-F1 across corpora.} Macro-F1 (mean $\pm$ std, 3 seeds) of frozen-encoder linear probes trained and tested within each domain. The handcrafted XGBoost baseline (rightmost group) beats WavLM and CLAP on DAC and ties AST ($0.271$ vs.\ $0.273$), with only wav2vec2 clearly ahead ($0.318$).}
\label{fig:b1}
\end{figure}

\textbf{In-domain results (C1).}
Table~\ref{tab:b1} and Figure~\ref{fig:b1} show within-domain probe performance. The same encoder varies by $0.57$--$0.80$ macro-F1 depending on the corpus: WavLM, for instance, drops from $1.000$ on BCSD-U to $0.205$ on DAC. AudioSet-pretrained encoders (CLAP, AST) dominate on the cleaner corpora --- an internal comparison, since prior evidence shows only that generic audio models underperform on non-speech vocalizations unless in-domain vocalization data is added \citep{gong2022vocalsound}; wav2vec2 is the only deep encoder that clearly wins on the noisy 8\,kHz DAC corpus, where the handcrafted XGBoost baseline ($0.271$) beats WavLM ($0.205$) and CLAP ($0.219$) and essentially ties AST ($0.273$): on the hardest domain, generic features compete with foundation encoders. C1 is supported, with two qualifications: BCSD-U is near a measurement ceiling ($n_{\mathrm{test}}=30$; even a perfect cell has a Clopper--Pearson lower bound of only $0.884$ single-seed / $0.960$ pooled, and WavLM now sits at pooled accuracy $0.967$, CP95 lower bound $0.906$; Table~\ref{tab:bcsduci}), and our comparison is internal — we do not re-run published per-corpus systems.

\begin{table}[t]
\centering
\footnotesize
\caption{In-domain macro-F1 of frozen encoders with linear probes (UAR in parentheses), mean over 3 seeds. Logistic-regression rows are near-deterministic: std $<0.005$ except CLAP on DAC ($0.0401$), residual nondeterminism from feature extraction; XGBoost std is $\pm0.0058/\pm0.0213$ on CC/DAC. Best deep encoder per column in bold. BCSD-U is near-ceiling ($n_{\mathrm{test}}=30$). A zero-shot MLLM baseline probed under deliberately asymmetric conditions is reported in Appendix~\ref{app:mllm} only.}
\label{tab:b1}
\setlength{\tabcolsep}{3.5pt}
\begin{tabular}{@{}lccc@{}}
\toprule
Encoder & BCSD-U & CC & DAC \\
\midrule
WavLM & \textbf{1.000} (1.000) & 0.740 (0.747) & 0.205 (0.255) \\
wav2vec2 & 0.883 (0.871) & 0.736 (0.738) & \textbf{0.318} (0.418) \\
CLAP & \textbf{1.000} (1.000) & \textbf{0.829} (0.824) & 0.219 (0.300) \\
AST & \textbf{1.000} (1.000) & 0.798 (0.802) & 0.273 (0.357) \\
\midrule
handcrafted+LR & 0.961 (0.944) & 0.816 (0.820) & 0.192 (0.258) \\
handcrafted+XGB & 0.961 (0.944) & 0.824 (0.816) & 0.271 (0.347) \\
\bottomrule
\end{tabular}
\end{table}

\textbf{Zero-shot MLLM baseline.}
A post-hoc zero-shot control with a general audio--language MLLM, run under deliberately asymmetric conditions, is reported in Appendix~\ref{app:mllm}; no claim in this paper depends on it.

\subsection{Within-Corpus Train--Test Leakage Audit}
\label{sec:within}

The cross-domain audit above cannot see leakage \emph{inside} one corpus, which is the path that inflates the diagonal --- the NTR denominator. We therefore audited each corpus's train/validation/test splits with exact hashing, a perceptual fingerprint (64-bit log-mel simhash), embedding nearest neighbors (wav2vec2 cosine), and waveform time-shift cross-correlation on the candidate pairs (details and per-method tables in Appendix~\ref{app:withinaudit}).

\textbf{What \texttt{group\_id} actually is.}
The group semantics differ per corpus and are weaker than ``infant'' or ``session'': in DAC, groups are the \texttt{uuid} filename prefix (uploader/device ID; same device, different epoch $=$ different recording); in BCSD-U, the group is the clip ID itself ($207$ clips / $207$ groups); in CC, groups merge only explicit \texttt{Uncom\_Rev\_} re-export pairs (median group size $1$). Group splits themselves are clean (no group crosses splits), but \emph{group-aware} throughout this paper means exactly ``aware of the groups recoverable from current metadata'' --- it cannot intercept same-recording or same-infant pairs that span groups, and we found such pairs in every corpus.

\textbf{Findings.}
No cross-split exact-hash hits exist, yet waveform-level matching does: CC has $42$ fingerprint hit pairs across train--test, of which $13$ are \emph{waveform-identical re-exports} (offset $0$, xcorr $\approx1.000$, different sha1 --- exact hashing cannot see them); BCSD-U has $5$ time-shifted slice pairs ($\pm1$--$2$\,s offsets) from the same long recording (\texttt{laugh\_1.m4a}) spanning train and test; DAC has $8$ fingerprint-only candidate pairs (single-method evidence, lower confidence). Removing the $65$ suspicious test clips ($8/13/44$ for DAC/BCSD-U/CC) and recomputing the full matrix under the unified protocol moves the diagonal by $\Delta\in[-0.072,+0.015]$ --- the largest drop is AST on DAC ($0.521\to0.449$, an 8-clip removal on a small test set), and three wav2vec2 cells move slightly \emph{up} --- so the diagonal is not systematically inflated. Cleaned NTR stays significantly positive for all five encoders ($0.198$--$0.337$ vs.\ $0.194$--$0.345$ before cleaning; Table~\ref{tab:withinnudtr}). We report the CC re-export pairs honestly as a construction flaw of the source distribution that byte-level auditing misses; the released manifest ships the exclusion list.

\subsection{Transfer Matrix and Few-Shot Frontier}
\label{sec:fewshot}

\begin{figure*}[t]
\centering
\includegraphics[width=0.62\textwidth]{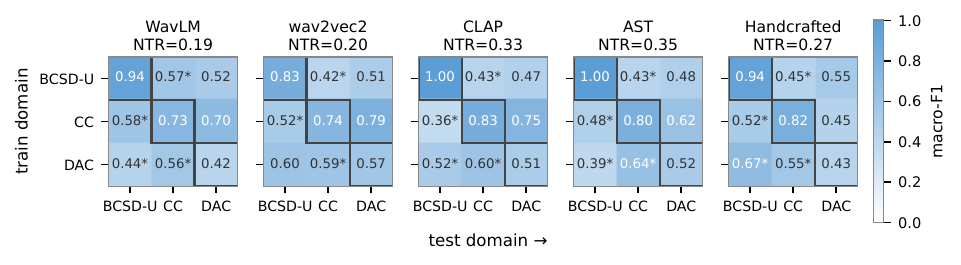}
\caption{\textbf{Cross-corpus transfer matrices (binary pain-vs-rest, unified held-out-test protocol).} Train $\times$ test macro-F1 for five encoders; diagonal cells (bold frame) exceed off-diagonal cells except in the DAC column (reverse transfer). $^*$: significant negative transfer (cluster bootstrap over \texttt{group\_id}, BH-FDR $q<0.05$). Panel titles give NTR (0.19--0.35).}
\label{fig:tmatrix}
\end{figure*}

\textbf{Transfer matrix (C2).}
Figure~\ref{fig:tmatrix} shows the binary \emph{pain-vs-rest} transfer matrices under the unified protocol of Section~\ref{sec:formulation} (every cell evaluated on the target domain's held-out test split), and Table~\ref{tab:ntr} summarizes them. NTR is positive for every encoder: $0.194$--$0.345$ on the full binary setting, rising to $0.239$--$0.461$ on the cry-only subset that removes non-cry clips (Figure~\ref{fig:ntr}): the transfer gap is not an artifact of non-cry material. Of the $30$ directed off-diagonal cells (5 encoders $\times$ 6 pairs), $19$ are raw-significant and $18$ show significant negative transfer after Benjamini--Hochberg FDR control at $q=0.05$ (paired cluster bootstrap over \texttt{group\_id}, $10{,}000$ resamples, recomputed on the corrected-$C$ predictions; wav2vec2 DAC$\to$BCSD-U is non-significant, $p=0.128$, and wav2vec2 DAC$\to$CC is marginal after FDR, adjusted $p=0.056$). On the cry-only subset, $21$ cells are raw-significant and $19$ survive FDR.
Transfer is asymmetric by up to $0.196$ F1 for the same corpus pair in opposite directions. C2 is supported. Relative to our original (superseded) protocol, the unified protocol lowers NTR for three of five encoders (by $0.046$--$0.075$) and raises it slightly for AST and handcrafted ($+0.020$/$+0.009$). All significance statements are group-level cluster bootstrap; clip-level resampling understates variance (per-cell flips in Appendix~\ref{app:protocol}).

\textbf{A systematic exception: reverse transfer into DAC.}
Every \emph{reversed} cell (off-diagonal above diagonal) involves DAC as the target, and the non-significant cells also cluster there. Training on the larger CC corpus ($n=541$ binary training clips) and testing on DAC beats DAC in-domain training for all five encoders in effect size, by $0.020$ (handcrafted) to $0.268$ (WavLM) macro-F1 --- but under the corrected trainer none of the five reverse cells is individually significant (cluster-bootstrap BH-adjusted $p\approx1.0$ throughout), so we report the exception as a consistent effect-size pattern, not a per-cell significant result. We read this as a data-mass effect, not a contradiction of C2: DAC's within-domain diagonal is unusually low (Table~\ref{tab:b1}), so a larger, more diverse training corpus acts as a regularizer that outweighs the domain shift. A size-matched control (CC subsampled to DAC's $n=315$) still beats the DAC diagonal for all three recomputed encoders (mean margin $+0.10$/$+0.20$/$+0.18$ for wav2vec2/CLAP/AST; all $9$ runs per encoder positive; the margin CI over the $3$ subsample draws includes zero for wav2vec2 and CLAP --- Table~\ref{tab:ccsub}), so size alone does not explain the exception, though the control is low-powered.

\begin{table}[t]
\centering
\footnotesize
\caption{Negative-transfer ratio per encoder under the unified held-out-test protocol (binary setting and cry-only subset) and the largest pairwise asymmetry $A(i,j)=|T[i,j]-T[j,i]|$. The handcrafted row uses the logistic-regression probe (as in the transfer runs), not the XGBoost variant of Table~\ref{tab:b1}.}
\label{tab:ntr}
\setlength{\tabcolsep}{2.5pt}
\begin{tabular}{@{}lccc@{}}
\toprule
Encoder & NTR (binary) & NTR (cry-only) & Max asymmetry \\
\midrule
WavLM & 0.194 & 0.256 & 0.134 (CC$\leftrightarrow$DAC) \\
wav2vec2 & 0.200 & 0.338 & 0.196 (CC$\leftrightarrow$DAC) \\
CLAP & 0.330 & 0.408 & 0.152 (CC$\leftrightarrow$DAC) \\
AST & 0.345 & 0.461 & 0.093 (DAC$\leftrightarrow$BCSD-U) \\
handcrafted & 0.271 & 0.239 & 0.127 (BCSD-U$\leftrightarrow$DAC) \\
\bottomrule
\end{tabular}
\end{table}

\textbf{Multi-level leakage audit.}
Byte-level deduplication cannot catch re-encoded or trimmed copies, so we audited embedding-level near-duplicates (CLAP cosine nearest neighbors, $\tau=0.98$, all six domain pairs). BCSD-U$\leftrightarrow$DAC is clean (max $0.94$); DAC$\leftrightarrow$CC, however, contains $157$ pairs above $0.98$ ($159$ pairs and $312$ clips in total, label agreement $99.4\%$), and the $15$ pairs above $0.99$ are $1{:}1$ matches confirmed in a second encoder --- re-encoded DAC copies inside CC that byte hashing cannot see (provenance evidence in Appendix~\ref{app:leakage}). Removing all $312$ involved clips leaves NTR positive for all three recomputed encoders (wav2vec2 $0.148$, CLAP $0.300$, AST $0.284$), and the CC$\to$DAC exception \emph{strengthens} in effect size ($0.778$--$0.821$ vs.\ clean DAC diagonal $0.442$--$0.448$), so reverse transfer is not a leakage artifact. Excluding instead the $346$ conflict-involved DAC clips leaves NTR at $0.34$--$0.41$ (DAC test $n=16$: low power; Appendix~\ref{app:leakage} only).

\textbf{Few-shot frontier (C3).}
Table~\ref{tab:fewshot} and Figure~\ref{fig:fewshot} show the label-budget sweep on the CC target domain with wav2vec2. The pre-specified fine-tuning arm B failed as a measurement instrument: at the full-data budget its per-seed macro-F1 is $0.7025/0.2727/0.3849$ (mean $0.453$, std $0.223$, far above the pre-specified std gate). This instability matches the predicted feature distortion of full fine-tuning under distribution shift \citep{kumar2022finetuning}; we therefore do not use Arm B for the verdict and fall back to the stabilized arm Bs, a clearly marked \emph{post-hoc} protocol fix.

Judged on Arms A, C, and Bs under the inductive DAPT protocol (target validation/test waveforms excluded; Appendix~\ref{app:protocol}), the low-label picture is saturation plus a low-budget DAPT advantage over fine-tuning, and the crossover is weaker than pre-specified: no arm separates at 1-shot (B $0.527$, A $0.517$, C $0.515$), and frozen probes match or exceed C at every budget up to 50-shot (episode-averaged A $\geq$ C). On paired episodes (shared shot draws, averaged over optimization seeds $42/43/44$), C significantly beats Bs at 5- and 10-shot (Bs$-$C: $-0.143$ [$-0.184$,$-0.103$] at 5-shot; $-0.125$ [$-0.172$,$-0.079$] at 10-shot), the 1-shot DAPT advantage is not robust to optimization-seed variance ($-0.052$; CI $[-0.124,+0.020]$ includes zero), and at 50-shot Bs exceeds C only as a \emph{trend} ($+0.022$; paired 95\% CI $[-0.009,+0.053]$ includes zero --- no significant difference), so the previously reported crossover does not hold up statistically; stabilized fine-tuning clearly wins only at the full-data budget ($0.822$).\footnote{Treating episode$\times$seed runs as $30$ paired observations instead of seed-averaging excludes zero at 1-shot (Bs$-$C CI $[-0.097,-0.006]$) and at 50-shot ($[+0.003,+0.041]$); we report the seed-averaged pairing as the conservative headline. Probe arms A/C are exactly deterministic in the optimization seed (scores identical across $42/43/44$), so their episode variance is pure shot-sampling noise.} Two asymmetries qualify all arm comparisons: Arm~C's DAPT stage consumes a fixed pool of $4{,}537$ unlabeled waveforms including target-domain (CC) train audio --- a fixed target-unlabeled exposure that the label-budget axis does not count, so the arms' total data budgets are intentionally unequal; and the episode protocol, originally single-seed, now covers three optimization seeds ($42/43/44$), with residual seed sensitivity at 1-shot reported above rather than modeled away. Episode-level std ($0.03$--$0.11$) dwarfs optimization variance --- the 1-shot arm differences of single-subset runs are within sampling noise (paired A$-$C CI $[-0.020,+0.109]$). C3 is therefore \emph{partially supported}: the full-budget ranking holds and DAPT helps at low budgets, but frozen probes saturate at low-to-moderate budgets, the 1-shot DAPT advantage does not survive the inductive, episode-averaged, multi-seed protocol, the 50-shot crossover is a non-significant trend, the high-label verdict rests on the post-hoc Bs arm, and the one-epoch DAPT budget is a lower bound (dose--response left to future work).

\begin{table}[t]
\centering
\footnotesize
\caption{Few-shot sweep on the CC target domain (wav2vec2, macro-F1, mean$\pm$std, 3 seeds, single fixed shot subset per budget). A: frozen probe; B: full fine-tuning (pre-specified, collapsed); C: DAPT+probe (inductive protocol: target validation/test waveforms excluded from DAPT); Bs: stabilized fine-tuning (post-hoc). Best per row in bold. Std$<0.001$ entries are \emph{deterministic} given the fixed shot subset; episode-averaged variants in Table~\ref{tab:episodes}. Arm~C values are from the authoritative inductive run (\texttt{B3\_fewshot\_agg\_inductive.csv}; C full $=0.705$); the superseded transductive run gave C full $=0.761$ (\texttt{B3\_fewshot\_agg.csv}) and is retained only for provenance in Table~\ref{tab:inductive}.}
\label{tab:fewshot}
\setlength{\tabcolsep}{3.2pt}
\begin{tabular}{@{}lcccc@{}}
\toprule
$n$-shot & A (probe) & B (full FT) & C (DAPT) & Bs (stab.\ FT) \\
\midrule
1 & 0.517{\scriptsize$\pm$.031} & \textbf{0.527}{\scriptsize$\pm$.096} & 0.515{\scriptsize$\pm$.061} & 0.413{\scriptsize$\pm$.202} \\
5 & 0.514{\scriptsize$\pm$.192} & \textbf{0.577}{\scriptsize$\pm$.091} & 0.528{\scriptsize$\pm$.076} & 0.436{\scriptsize$\pm$.152} \\
10 & 0.679{\scriptsize$\pm$.014} & \textbf{0.688}{\scriptsize$\pm$.017} & 0.655{\scriptsize$\pm$.053} & 0.615{\scriptsize$\pm$.054} \\
50 & 0.676{\scriptsize$\pm$.035} & 0.594{\scriptsize$\pm$.183} & 0.637{\scriptsize$\pm$.011} & \textbf{0.684}{\scriptsize$\pm$.032} \\
full & 0.736{\scriptsize$\pm$.000} & 0.453{\scriptsize$\pm$.223} & 0.705{\scriptsize$\pm$.000} & \textbf{0.822}{\scriptsize$\pm$.022} \\
\bottomrule
\end{tabular}
\end{table}

\subsection{Ontology Joint Training and Quality Analysis}
\label{sec:joint}

\begin{figure}[t]
\centering
\includegraphics[width=0.71\columnwidth]{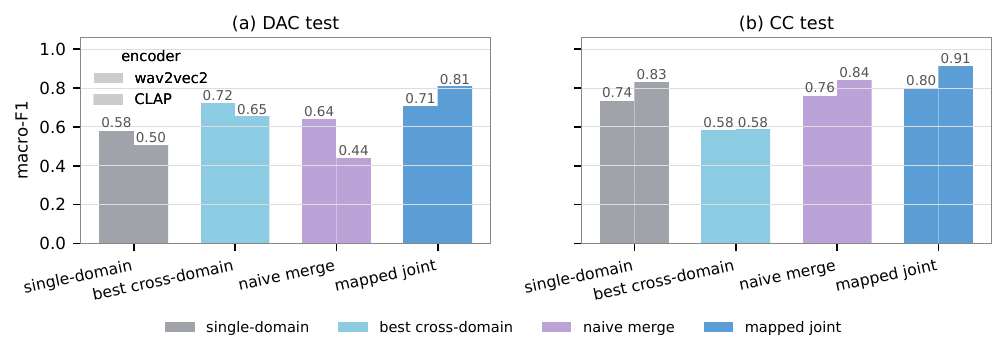}
\caption{\textbf{Ontology-mapped joint training beats single-domain, best cross-domain, and naive merging.} Macro-F1 on the DAC test set (a) and CC test set (b). Merging raw, unmapped label spaces degrades DAC-test performance below the best cross-domain transfer, directly evidencing the cost of label-ontology mismatch.}
\label{fig:joint}
\end{figure}

\textbf{Mapped joint training (C4).}
Table~\ref{tab:joint} and Figure~\ref{fig:joint} compare four training conditions on the two-class \emph{hunger}/\emph{pain} evaluation aligned with the transfer matrix. Mapped joint training is best in all four encoder$\times$target combinations. The naive merge condition — pooling raw labels without the ontology (construction in Appendix~\ref{app:naive}) — quantifies the cost of label-space mismatch: it falls below the best \emph{cross-domain} transfer on DAC-test ($0.639$ vs.\ $0.786$ for wav2vec2; $0.435$ vs.\ $0.749$ for CLAP) and loses $0.039$--$0.371$ F1 to mapped joint training. We scope the claim accordingly: \emph{in the tested binary, shared-label settings} (two encoders, two target domains), explicit ontology mapping beats naive merging --- we do not claim the five-class ontology is validated in full. Two further caveats: the mappings are author-set rules without independent double annotation, disagreement rates, or blind review; and we did not run intermediate baselines (label-name normalization without the full ontology) or mapping-sensitivity analyses (leave-one-mapping-out, alternative mappings), so the margin attributable to the ontology \emph{per se} versus to any sensible normalization is unknown (Limitations). A small-corpus case study on the 46-clip DI-$\Delta$ is positive but underpowered ($n=2$ pain clips) and co-occurs with degradation on a small DAC \emph{soothe}/\emph{pain} subset ($n=12$; e.g., $1.000\to0.429$); we report it as a directional case study only (Appendix~\ref{app:didelta}); we do not claim small corpora gain most.

\textbf{Task-definition check.}
The binary \emph{pain-vs-rest} matrix pools different ``rest'' classes per domain, so part of the measured gap could reflect task-definition differences rather than acoustics. For DAC$\leftrightarrow$CC, which share \emph{hunger}/\emph{pain}, the common-label matrix (Table~\ref{tab:commonlabel}) gives mean absolute gap $0.176$ vs.\ $0.182$ binary: task definition accounts for at most $\sim$3\% of the gap, and the sign pattern — including CC$\to$DAC reverse transfer — is unchanged. The DAC$\leftrightarrow$BCSD-U common-label matrix is underpowered ($n_{\mathrm{test}}=11$--$12$; mean $|$gap$|$ $0.565$ vs.\ $0.245$ binary; appendix only).

\begin{table}[t]
\centering
\footnotesize
\caption{Joint training under the ontology (macro-F1; LogReg arm, deterministic, std$<0.001$). All rows use the same unified held-out-test protocol: single-domain and best cross-domain are taken from the two-class transfer runs on the target test split; naive merge pools raw labels; mapped joint pools labels through $\mathcal{O}$. Best per column in bold.}
\label{tab:joint}
\setlength{\tabcolsep}{3.2pt}
\begin{tabular}{@{}lcccc@{}}
\toprule
& \multicolumn{2}{c}{wav2vec2} & \multicolumn{2}{c}{CLAP} \\
\cmidrule(lr){2-3}\cmidrule(l){4-5}
Condition & DAC & CC & DAC & CC \\
\midrule
Single-domain & 0.571 & 0.736 & 0.508 & 0.829 \\
Best cross-domain & 0.786 & 0.590 & 0.749 & 0.597 \\
Naive merge & 0.639 & 0.761 & 0.435 & 0.840 \\
Mapped joint & \textbf{0.706} & \textbf{0.800} & \textbf{0.806} & \textbf{0.911} \\
\bottomrule
\end{tabular}
\end{table}

\textbf{Quality--transfer correlation (C5).}
\label{sec:quality}
Figure~\ref{fig:quality} pools the point-biserial correlation between clip-level SNR and cross-domain correctness over the six directed domain pairs per encoder. The mean correlation is significantly \emph{negative} for all encoders ($-0.04$ to $-0.10$; Fisher combined $p<0.01$; $21/30$ pairs negative) — the opposite of the quality hypothesis. Two validity problems disqualify the pre-specified reading: the SNR estimator is not comparable across sampling rates (BCSD-U's $102.9$\,dB is a saturation artifact on clean 44.1\,kHz audio, not a physical SNR), and SNR is almost perfectly confounded with domain identity — the high-``SNR'' domain is simply the hardest transfer target (incoming cross-domain F1 $0.496$ vs.\ $0.531$/$0.524$). Within-domain correlations are at most weak (Table~\ref{tab:withinsnr}: none significant on DAC; two of five positive and significant on CC). C5 is \emph{refuted as stated}: the pooled correlation is a domain-identity artifact --- corpus identity, not clip-level signal quality, dominates cross-corpus measurements \citep{braunschweiler2021corpus}.

%% file: sections/conclusion.tex
\section{Conclusion}
\label{sec:conclusion}

We audited how far off-the-shelf audio foundation models transfer to infant cry across four public corpora, under a unified five-class need ontology with a multi-level leakage audit. The answers are measurable and mostly uncomfortable: within-domain performance swings by $0.57$--$0.80$ macro-F1 across corpora, and a zero-shot general audio MLLM trails even the weakest in-domain linear probe (C1; Appendix-only control); negative transfer is the norm (NTR $0.19$--$0.35$; $0.24$--$0.46$ cry-only), asymmetric, and significant in $18$ of $30$ directed cells after FDR control (cluster bootstrap over recovered-metadata groups), with one systematic exception — reverse transfer \emph{into} the noisy, low-diagonal DAC corpus, consistent in effect size though not individually significant — that survives near-duplicate removal (C2); frozen probes saturate at modest label budgets, DAPT significantly beats stabilized fine-tuning at $5$--$10$-shot (the 1-shot advantage is not robust to optimization-seed variance), the 50-shot crossover is a non-significant trend (seed-averaged paired CI $[-0.009,+0.053]$) that materializes only at the full-data budget, the 1-shot arm differences are within episode-level sampling noise, and naive full fine-tuning collapses outright (C3, partially supported); in the tested binary, shared-label settings, ontology-mapped joint training is best in every encoder$\times$target setting, while merging unmapped labels can be worse than training cross-domain without merging (C4); a pre-specified quality--transfer correlation is refuted as stated, the pooled effect being a domain-identity artifact (C5); and $349$ content-identical clip groups in the raw corpora carry conflicting metadata labels across distributions, $292$ of them ($83.7\%$) disputes over \emph{hungry} (C6). Task-definition differences explain at most $\sim$3\% of the DAC$\leftrightarrow$CC gap (common-label re-computation). Removal robustness is scoped to what we actually recomputed: \emph{the core C2 NTR sign and the CC$\to$DAC exception survive removal of the $157$ near-duplicate pairs for the recomputed encoders} (wav2vec2, CLAP, AST); C1, C3, C4, and C5 were not re-run under removal. A within-corpus train--test audit further found cross-group leakage the group split cannot intercept (13 waveform-identical CC re-export pairs, 5 shifted-slice BCSD-U pairs, 8 fingerprint-only DAC pairs); after removing the $65$ suspicious test clips the diagonal moves by at most $-0.072$ and cleaned NTR stays significantly positive for all encoders ($0.198$--$0.337$).

\textbf{Limitations.}
Several cells are weakly powered: DI-$\Delta$ has $n=2$ pain-discomfort clips, so the small-corpus case study is directional; BCSD-U sits near a measurement ceiling ($n_{\mathrm{test}}=30$, pooled CP95 lower bound $0.906$--$0.960$), inflating absolute gaps and NTR; the conflict-excluded DAC test split ($n=16$) and the DAC$\leftrightarrow$BCSD-U common-label matrix are underpowered; the one-epoch DAPT budget makes Arm C a lower bound and leaves dose--response open; deterministic probe rows report std$\approx0$ that reflects protocol determinism, not robustness; the zero-shot MLLM comparison (Appendix~\ref{app:mllm}) is deliberately asymmetric (60-clip subsamples on CC/DAC; API versioning outside our control, snapshot 2026-07-21); and need labels are caregiver-annotated conventions, not physiological ground truth. The ontology mappings are author-set rules: no independent double annotation, no label-name-normalization intermediate baseline, and no leave-one-mapping-out sensitivity analysis were run, so the C4 margin cannot be attributed to the ontology beyond ``explicit mapping beats naive merge in the tested binary settings''. The few-shot episode protocol now covers three optimization seeds ($42/43/44$) --- probe arms are exactly deterministic in the seed and the 1-shot arm ordering is pairing-sensitive --- and Arm~C comparisons involve an intentionally asymmetric, fixed target-unlabeled data budget. \texttt{group\_id} semantics are per-corpus metadata proxies (device prefix, clip ID, or re-export pairs), not infant identity: the within-corpus audit (Section~\ref{sec:within}) found cross-group leakage the group split cannot intercept, and residual same-infant leakage beyond what four matching methods recover cannot be excluded. The 349-conflict audit mixes provenance types --- independent re-annotation, repackaging, and rule-mapping conflicts cannot be fully separated --- so we report them as cross-distribution metadata/label conflicts, not as 349 independent caregiver re-annotations. Removal robustness covers the recomputed encoders' C2/NTR and CC$\to$DAC cells only. Two corpora (BCSD-U, CC) redistribute without an explicit license and their original distribution channels could not be fully verified; they are used for evaluation only, no audio is redistributed, and anonymous-release reproducibility is layered (Appendix~\ref{app:repro}).

\textbf{Outlook.}
The transfer gap, label conflicts, and reverse-transfer exception point the same way: fewer cry classifiers, more shared infrastructure — pooled unlabeled cry audio, explicit ontologies, leakage screening, unified evaluation. We release the mapping rules, manifests, and code.

%% file: sections/appendix.tex
\section{Corpus Inventory and Licenses}
\label{app:corpora}

Table~\ref{tab:corpora} (main text) summarizes the audited corpora; Table~\ref{tab:appcorpora} extends it with provenance. Dedup removes $1{,}247$ clips before splitting (BCSD $781$, DeepInfant $466$). BCSD's \emph{hungry} subset is byte-identical to DAC's and is removed from BCSD-U; DeepInfant V1 is a repackaging of DAC, so only the V2 increment survives as DI-$\Delta$. Two corpora redistribute without an explicit license file (BCSD-U, CC), and their original distribution channels could not be fully verified; we therefore use them for evaluation only and release no audio. CryCeleb \citep{budaghyan2024cryceleb} is used exclusively as unlabeled DAPT audio under the license terms as stated by its distributors (CC BY-NC-ND 4.0, research use only); we release no CryCeleb audio, no derivative clips, and no derivative model weights trained on it.

\begin{table}[!htb]
\centering
\small
\caption{Full corpus audit. ``Mapped'' is the clip count after deduplication and ontology mapping (excluding $\bot$ labels).}
\label{tab:appcorpora}
\setlength{\tabcolsep}{4pt}
\begin{tabular}{@{}lccccc@{}}
\toprule
Corpus & Scanned & Removed & Mapped & Rate & License \\
\midrule
BCSD-U & 988 & 781 & 207 & 44.1\,k & unspecified \\
CC & 786 & 0 & 786 & 8\,k & unspecified \\
DAC & 457 & 0 & 457 & 8\,k & ODbL-1.0 \\
DI-$\Delta$ & 597 & 466 & 46 & 44.1\,k & Apache-2.0 \\
\midrule
Total & --- & 1247 & 1496 & --- & --- \\
\bottomrule
\end{tabular}
\end{table}

The group-stratified split is train/validation/test $=1037/227/232$ clips; groups (recovered-metadata groups --- per-corpus semantics audited in Appendix~\ref{app:withinaudit}) never cross splits.

\section{Ontology Mapping and Conflict Audit}
\label{app:ontology}

Table~\ref{tab:appmapping} gives the full raw-label-to-ontology mapping; Table~\ref{tab:appconflicts} lists the largest conflict groups.

\begin{table}[!htb]
\centering
\small
\caption{Full ontology mapping. $\bot$: excluded. *: file removed if byte-identical to a DAC file.}
\label{tab:appmapping}
\setlength{\tabcolsep}{4pt}
\begin{tabular}{@{}lllll@{}}
\toprule
Ontology & DAC & DI-$\Delta$ & BCSD-U & CC \\
\midrule
hunger & hungry & hungry* & hungry* & Hungry \\
pain-disc. & belly\_pain & belly\_pain & belly pain & Pain \\
 & discomfort & discomfort & discomfort & Uncomf. \\
 &  & cold\_hot & cold\_hot &  \\
sleepiness & tired & tired & --- & --- \\
need-sooth. & burping & burping & burping & --- \\
 &  & lonely &  &  \\
 &  & scared &  &  \\
neutral & --- & --- & laugh & --- \\
 &  &  & noise &  \\
$\bot$ & --- & unknown & --- & --- \\
\bottomrule
\end{tabular}
\end{table}

\begin{table}[!htb]
\centering
\small
\caption{Top cross-corpus label-conflict pairs among the 349 sha1-identical groups (computed on the raw, pre-deduplication corpus collection).}
\label{tab:appconflicts}
\setlength{\tabcolsep}{4pt}
\begin{tabular}{@{}lcc@{}}
\toprule
Label pair & Groups & Files \\
\midrule
discomfort $\mid$ hungry & 87 & 348 \\
cold\_hot $\mid$ hungry & 86 & 344 \\
belly\_pain $\mid$ hungry & 61 & 244 \\
burping $\mid$ hungry & 57 & 228 \\
cold\_hot $\mid$ discomfort & 14 & 56 \\
belly\_pain $\mid$ discomfort & 13 & 49 \\
\bottomrule
\end{tabular}
\end{table}

A recomputation of conflicts from the raw corpora yields $349$ groups (an earlier reference table shipped with the corpora lists $342$); we report the recomputed number throughout. We classify these as \emph{cross-distribution metadata/label conflicts}, not as $349$ independent caregiver re-annotations: by provenance, part plausibly reflect independent re-annotation of shared source recordings, part dataset repackaging or directory-naming conventions (the BCSD \emph{hungry} set is byte-identical to DAC's), and part secondary rule mapping; for a substantial fraction the provenance cannot be determined from the released files, so we do not report a per-source breakdown and treat the count as an upper bound on genuinely independent re-annotation. Within the released (post-dedup) manifest, all $341$ sha1 keys and $8$ filename keys of these groups are present exactly once, with the retained copy on the DAC side ($346$ clips: $234$ train, $50$ val, $62$ test); cross-domain byte-identical duplicates in the manifest are zero.

\section{Leakage Audit Details}
\label{app:leakage}

\textbf{Embedding-level near-duplicate audit.}
We computed CLAP embeddings for all $1{,}496$ manifest clips and, for each ordered pair of domains, the cross-domain cosine similarity of every clip pair. Table~\ref{tab:neardupaudit} summarizes the maxima and exceedance counts; Table~\ref{tab:neardup} reports the removal-robustness of the transfer matrix. The $15$ DAC$\leftrightarrow$CC pairs above cosine $0.99$ are $1{:}1$, label-consistent matches, independently confirmed in wav2vec2 space (cosine $\geq0.926$); the CC-side files carry \texttt{orig\_sr=8000} (phone-recording provenance like DAC) while differing in sha1 — i.e., transcoded or trimmed DAC copies inside CC.

\begin{table}[!htb]
\centering
\footnotesize
\caption{Cross-domain embedding near-duplicate audit (CLAP cosine). $\tau=0.98$ defines suspicious pairs.}
\label{tab:neardupaudit}
\setlength{\tabcolsep}{3.5pt}
\begin{tabular}{@{}lcccc@{}}
\toprule
Domain pair & max cos & $>0.95$ & $>0.98$ & $>0.99$ \\
\midrule
DAC $\mid$ BCSD-U & 0.942 & 0 & 0 & 0 \\
DAC $\mid$ CC & 0.994 & 1592 & \textbf{157} & \textbf{15} \\
DAC $\mid$ DI-$\Delta$ & 0.987 & 54 & 2 & 0 \\
BCSD-U $\mid$ CC & 0.962 & 9 & 0 & 0 \\
BCSD-U $\mid$ DI-$\Delta$ & 0.913 & 0 & 0 & 0 \\
CC $\mid$ DI-$\Delta$ & 0.969 & 30 & 0 & 0 \\
\bottomrule
\end{tabular}
\end{table}

\begin{table}[!htb]
\centering
\footnotesize
\caption{Transfer-matrix robustness after removing all $312$ clips involved in suspicious near-duplicate pairs (unified protocol; all cells after exclusion). The CC$\to$DAC reverse-transfer exception \emph{strengthens} in effect size after removal, so it is not a leakage artifact; part of the strengthening reflects the DAC diagonal dropping as its training set shrinks. The recomputation covers three of the five encoders (wav2vec2, CLAP, AST); WavLM and the handcrafted baseline were omitted under the compute budget of the audit window. Robustness claims in the main text are scoped accordingly: only the C2 NTR sign and the CC$\to$DAC exception are claimed to survive removal, for these recomputed encoders.}
\label{tab:neardup}
\setlength{\tabcolsep}{2.5pt}
\begin{tabular}{@{}lcccc@{}}
\toprule
Encoder & NTR (full) & NTR (excl.) & CC$\to$DAC & DAC diag \\
\midrule
wav2vec2 & 0.200 & 0.148 & 0.803 & 0.442 \\
CLAP & 0.330 & 0.300 & 0.778 & 0.448 \\
AST & 0.345 & 0.284 & 0.821 & 0.442 \\
\bottomrule
\end{tabular}
\end{table}

\textbf{Conflict-exclusion sensitivity.}
Excluding all $346$ DAC clips involved in the pre-dedup conflict audit (DAC splits shrink $315/64/78 \to 81/14/16$) leaves NTR positive and numerically larger (Table~\ref{tab:conflictrobust}). Because the residual DAC test split has only $n=16$ clips, this variant is underpowered; we report it as a sensitivity analysis only.

\begin{table}[!htb]
\centering
\footnotesize
\caption{NTR before/after excluding the $346$ conflict-involved DAC clips (unified protocol; DAC test $n=16$ after exclusion — low power). DAC diag: true DAC diagonal macro-F1; note that for CLAP and AST the diagonal \emph{drops} after exclusion, while for wav2vec2 it rises.}
\label{tab:conflictrobust}
\setlength{\tabcolsep}{2.5pt}
\begin{tabular}{@{}lccc@{}}
\toprule
Encoder & NTR (full) & NTR (excl.) & DAC diag (full$\to$excl.) \\
\midrule
wav2vec2 & 0.200 & 0.413 & 0.571 $\to$ 0.816 \\
CLAP & 0.330 & 0.340 & 0.508 $\to$ 0.429 \\
AST & 0.345 & 0.347 & 0.521 $\to$ 0.467 \\
\bottomrule
\end{tabular}
\end{table}

\section{Multi-Class and Common-Label Matrices}
\label{app:mc}

Two-class matrices restricted to \emph{hunger}/\emph{pain} (CC$\leftrightarrow$DAC) reproduce the binary pattern of Figure~\ref{fig:tmatrix}; Table~\ref{tab:clsummary} summarizes the gap comparison and Table~\ref{tab:commonlabel} gives the full per-encoder cells. A \emph{need-soothing}/\emph{pain} matrix (DAC$\leftrightarrow$BCSD-U) includes cells with $n_{\mathrm{test}}=11$--$12$ clips; its mean $|$gap$|$ is $0.565$ (vs.\ $0.245$ binary), but at this sample size the numbers are not interpretable, so we use them only qualitatively and mark them underpowered. Full per-cell numbers are released with the code.

\begin{table}[!htb]
\centering
\small
\caption{Task-definition check: mean $|$gap$|$ (diag $-$ off-diag, absolute) over the two directed DAC$\leftrightarrow$CC cells, binary \emph{pain-vs-rest} vs.\ common-label \emph{hunger}/\emph{pain} protocol.}
\label{tab:clsummary}
\setlength{\tabcolsep}{4pt}
\begin{tabular}{@{}lcc@{}}
\toprule
Encoder & binary & common-label \\
\midrule
WavLM & 0.221 & 0.242 \\
wav2vec2 & 0.180 & 0.175 \\
CLAP & 0.236 & 0.239 \\
AST & 0.132 & 0.078 \\
handcrafted & 0.140 & 0.147 \\
\midrule
mean & 0.182 & 0.176 \\
\bottomrule
\end{tabular}
\end{table}

\begin{table}[!htb]
\centering
\small
\caption{Common-label vs.\ binary transfer cells for DAC$\leftrightarrow$CC (macro-F1, unified protocol; deterministic, std$<0.001$). Binary: \emph{pain-vs-rest}. Common-label: restricted to \emph{hunger}/\emph{pain} in both domains. The gap pattern is preserved; the CC$\to$DAC reverse-transfer exception holds under both protocols.}
\label{tab:commonlabel}
\setlength{\tabcolsep}{3.5pt}
\begin{tabular}{@{}lcccc@{}}
\toprule
& \multicolumn{2}{c}{binary} & \multicolumn{2}{c}{common-label} \\
\cmidrule(lr){2-3}\cmidrule(l){4-5}
Cell & WavLM & wav2vec2 & WavLM & wav2vec2 \\
\midrule
DAC diag & 0.422 & 0.571 & 0.417 & 0.580 \\
CC diag & 0.730 & 0.736 & 0.730 & 0.736 \\
DAC$\to$CC & 0.561 & 0.590 & 0.549 & 0.608 \\
CC$\to$DAC & 0.695 & 0.786 & 0.720 & 0.801 \\
\midrule
& \multicolumn{2}{c}{binary} & \multicolumn{2}{c}{common-label} \\
\cmidrule(lr){2-3}\cmidrule(l){4-5}
& CLAP & AST & CLAP & AST \\
\midrule
DAC diag & 0.508 & 0.521 & 0.503 & 0.595 \\
CC diag & 0.829 & 0.798 & 0.829 & 0.798 \\
DAC$\to$CC & 0.597 & 0.637 & 0.615 & 0.668 \\
CC$\to$DAC & 0.749 & 0.623 & 0.767 & 0.621 \\
\midrule
& \multicolumn{4}{c}{handcrafted} \\
\midrule
DAC diag & 0.435 & --- & 0.431 & --- \\
CC diag & 0.816 & --- & 0.816 & --- \\
DAC$\to$CC & 0.555 & --- & 0.542 & --- \\
CC$\to$DAC & 0.455 & --- & 0.452 & --- \\
\bottomrule
\end{tabular}
\end{table}

\begin{figure}[t]
\centering
\includegraphics[width=\columnwidth]{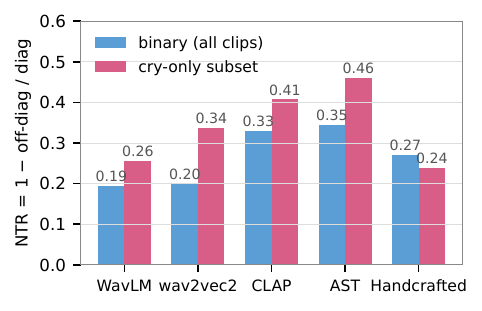}
\caption{\textbf{NTR is positive for all encoders and grows on the cry-only subset.} NTR under the unified held-out-test protocol on the full binary setting (blue) and on the cry-only subset (pink). Cry-only NTR rises to 0.24--0.46, showing the transfer gap is not an artifact of non-cry material.}
\label{fig:ntr}
\end{figure}

\section{Naive Merge Construction}
\label{app:naive}

The naive-merge baseline of Table~\ref{tab:joint} is constructed in three steps, using no ontology information during training:
\begin{enumerate}
\item \textbf{POOL}: concatenate the train splits of all source domains that share the target label pair (DAC$+$CC, clips whose ontology label is in \{\emph{hunger}, \emph{pain-discomfort}\}; $n=288+541=829$), keeping each clip's \emph{original raw label} as a domain-qualified class ($y=\text{domain}{::}\text{raw\_label}$, e.g.\ \texttt{DAC::hungry} vs.\ \texttt{CC::Hungry}; $6$ distinct raw classes).
\item \textbf{TRAIN}: fit the same StandardScaler$+$LogReg (lbfgs, class-weight balanced) probe on the pooled raw-labeled training set, selecting $C\in\{0.01,0.1,1,10\}$ on the pooled raw-labeled validation split (DAC val $+$ CC val) by macro-F1 over the raw classes.
\item \textbf{SCORE}: predict on each target domain's test split, map every predicted raw class back to its ontology label via the released raw$\to$ontology table, and report macro-F1/UAR against the ontology labels. Ontology information is used only during scoring, never during training.
\end{enumerate}

\section{DI-$\Delta$ Small-Corpus Case Study}
\label{app:didelta}

On the 46-clip DI-$\Delta$ corpus (\emph{need-soothing}/\emph{pain-discomfort}), adding mapped DI-$\Delta$ data to DAC training raises DI-$\Delta$ performance from $0.143$ to $0.894$ (wav2vec2) and from $0.417$ to $0.681$ (CLAP). Two qualifications apply, and we therefore present this as a directional case study rather than a claim: DI-$\Delta$ contains only $n=2$ pain-discomfort clips, so these numbers are underpowered; and performance on a small DAC \emph{soothe}/\emph{pain} subset ($n=12$) degrades under the same joint training (e.g., $1.000\to0.455$ for wav2vec2, $0.400\to0.333$ for CLAP), so the benefit is not uniform across evaluation slices. The two DAC-test protocols of Table~\ref{tab:joint} ($n=74$, \emph{hunger}/\emph{pain}) and this case study ($n=12$, \emph{soothe}/\emph{pain}) are different evaluation slices and should not be conflated.

\section{Few-Shot Details}
\label{app:fewshot}

\textbf{Arm B collapse record.} Full fine-tuning at the CC full-data budget gives per-seed macro-F1 $0.7025/0.2727/0.3849$ (mean $0.453$, std $0.223$); one seed fails already at 50-shot ($0.385$). On the DAC target the full-budget scores are A $0.580$, B $0.456$ (collapsed), C $0.520$ (inductive); no Bs arm was run for DAC.

\textbf{DAPT log.} One epoch of masked-prediction continuation on the pooled unlabeled cry audio (inductive pool: train splits $1{,}037$ clips $+$ CryCeleb $3{,}500$ clips $=4{,}537$; Appendix~\ref{app:protocol}): 283 steps, 96\,s wall time on one RTX 3090, final MLM loss $\approx1164$ (the loss magnitude reflects the raw-scale objective, not divergence). The budget is intentionally light; Arm C should be read as a lower bound, and the DAPT dose--response (whether a heavier budget widens or closes the low-label margin) is future work.

\textbf{Size-matched CC subsample (reverse-transfer robustness).}
To test whether the CC$\to$DAC exception is merely a data-mass effect, we subsampled CC training clips to the size of DAC's binary training set ($n=315$) and re-ran the transfer (9 runs per encoder: 3 subsamples $\times$ 3 seeds, seeds deterministic). Table~\ref{tab:ccsub} shows that size-matched CC training still beats DAC in-domain training in $9/9$ runs for all three recomputed encoders (paired margin mean $+0.099$/$+0.200$/$+0.181$ for wav2vec2/CLAP/AST), so the exception is not explained by training-set size alone; the dependency structure is 3 subsample draws $\times$ deterministic seeds, so the subsample-level CIs (Table~\ref{tab:ccsub}) are the honest uncertainty statement and include zero for two of the three encoders.

\begin{table}[!htb]
\centering
\footnotesize
\caption{CC$\to$DAC with CC subsampled to DAC's training size ($n=315$). Margin: subsampled CC$\to$DAC minus DAC in-domain (mean over 3 subsample draws; seeds are deterministic). All $9/9$ runs per encoder beat the DAC diagonal (paired margin $>0$ in every run); across only $3$ subsample draws the 95\% CI of the margin includes zero for wav2vec2 and CLAP ($[-0.12,+0.32]$ / $[-0.01,+0.41]$) but not AST ($[+0.05,+0.31]$), so the control is directionally consistent but low-powered.}
\label{tab:ccsub}
\setlength{\tabcolsep}{2.5pt}
\resizebox{\columnwidth}{!}{%
\begin{tabular}{@{}lcccc@{}}
\toprule
Encoder & DAC in-domain & full CC$\to$DAC & sub-315 CC$\to$DAC & margin \\
\midrule
wav2vec2 & 0.571 & 0.786 & $0.670\pm0.076$ & $+0.099$ \\
CLAP & 0.508 & 0.749 & $0.708\pm0.074$ & $+0.200$ \\
AST & 0.521 & 0.623 & $0.702\pm0.044$ & $+0.181$ \\
\bottomrule
\end{tabular}}
\end{table}

\begin{figure}[t]
\centering
\includegraphics[width=\columnwidth]{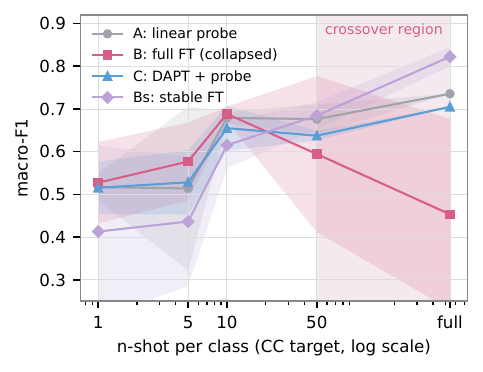}
\caption{\textbf{Few-shot adaptation curves on the CC target domain.} Macro-F1 (mean $\pm$ std, 3 seeds) vs.\ target labels per class (log axis; ``full'' $=541$ training clips). Arm C uses the inductive DAPT checkpoint (Appendix~\ref{app:protocol}); no arm separates at 1-shot, and B collapses with high variance at 50-shot/full. The stabilized fine-tuning arm (Bs) crosses A and C between 10-shot and 50-shot (shaded region) and reaches the best full-data score (0.822).}
\label{fig:fewshot}
\end{figure}

\section{Protocol Hardening: Inductive DAPT, Cluster Bootstrap, Episode Variance}
\label{app:protocol}

This section documents three protocol fixes applied after an external audit of our pre-specified analysis plan, and their effect on the reported numbers.

\textbf{Inductive DAPT re-run.}
Our first Arm~C run continued pretraining on the pooled unlabeled portions of \emph{all} splits of the four corpora plus CryCeleb --- transductive, in that target validation/test waveforms (unlabeled) entered DAPT. We re-ran DAPT with those waveforms excluded (train split $1{,}037$ clips $+$ the identical $3{,}500$ CryCeleb clips, same seed and ordering; $4{,}537$ clips, one epoch, $283$ steps, $96$\,s) and re-ran the full Arm~C sweep with the inductive checkpoint. Table~\ref{tab:inductive} compares the two protocols on the CC target. Under the inductive protocol Arm~C no longer leads at 1-shot; the Bs--C crossover region (10--50 shot) and the full-budget ranking Bs $>$ C $>$ A are unchanged. All Arm~C numbers in the main text use the inductive checkpoint; the transductive values are retained here for provenance only.

\begin{table}[!htb]
\centering
\footnotesize
\caption{Arm~C on the CC target, transductive vs.\ inductive DAPT (macro-F1, mean over 3 seeds, corrected $C$ selector). The transductive column is superseded and shown only for provenance. The 1-shot lead disappears under the inductive protocol; note that at the full budget the inductive Arm~C ($0.705$) no longer exceeds Arm~A ($0.736$).}
\label{tab:inductive}
\setlength{\tabcolsep}{3.5pt}
\begin{tabular}{@{}lccc@{}}
\toprule
$n$-shot & transductive (old) & inductive (new) & $\Delta$ \\
\midrule
1 & 0.536 & 0.515 & $-0.021$ \\
5 & 0.526 & 0.528 & $+0.002$ \\
10 & 0.618 & 0.655 & $+0.037$ \\
50 & 0.653 & 0.637 & $-0.016$ \\
full & 0.761 & 0.705 & $-0.056$ \\
\bottomrule
\end{tabular}
\end{table}

\textbf{Cluster bootstrap.}
Our first significance tests resampled clips; because the $1{,}496$ clips form only $1{,}160$ groups (infant/session/augmentation clusters of up to $15$ clips), clip-level resampling understates variance. We re-ran every directed-cell test as a cluster bootstrap over \texttt{group\_id} ($10{,}000$ resamples, same seed, BH-FDR at $q=0.05$ within each 30-cell setting). The counts below are all recomputed on the corrected-$C$ predictions (\texttt{run\_r1\_cluster.py} over \texttt{R1\_unified\_eval} and post-fix \texttt{B2\_predictions.csv}): the binary setting stands at $19/30$ raw-significant and $18/30$ after FDR (wav2vec2 DAC$\to$BCSD-U moved out of significance, $p=0.128$; wav2vec2 DAC$\to$CC is raw-significant at $p=0.036$ but marginal after FDR, adjusted $p=0.056$); the cry-only setting contracted from $22/30$ raw-significant under clip-level bootstrap to $21/30$ under cluster bootstrap, of which $19/30$ survive FDR. Exactly one cell flipped from raw-significant to non-significant between the clip-level and cluster resampling schemes (Table~\ref{tab:bootflip}); no cell flipped in the reverse direction, and per-cell effect sizes moved by at most $0.03$ (binary $0.026$, cry-only $0.021$). Under the corrected trainer the CC$\to$DAC reverse-transfer exception remains positive in effect size for all five encoders but is not individually significant (BH-adjusted $p\approx1.0$).

\begin{table}[!htb]
\centering
\footnotesize
\caption{Corrected-$C$ provenance: the single cell that flipped from raw-significant ($p<0.05$) under clip-level bootstrap to non-significant under cluster bootstrap (no reverse flips; binary setting: zero flips). Counts quoted in the main text are from the same corrected-$C$ rerun ($18/30$ binary and $19/30$ cry-only after BH-FDR).}
\label{tab:bootflip}
\setlength{\tabcolsep}{3pt}
\begin{tabular}{@{}llcc@{}}
\toprule
Setting & Cell & $p$ (clip) & $p$ (cluster) \\
\midrule
cry-only & CLAP BCSD-U$\to$DAC & $0.004$ & $0.092$ \\
\bottomrule
\end{tabular}
\end{table}

\textbf{Episode-level shot variance.}
The main-text sweep draws one shot subset per budget, shared across arms and seeds, so low-budget std reflects fitting stochasticity on a fixed subset only. We re-ran the CC sweep with an independent stratified shot draw per episode: $20$ episodes for Arms A/C, $10$ for Arm~Bs (GPU budget), each across three optimization seeds ($42/43/44$). Arms A/C are exactly deterministic in the optimization seed (scores identical across $42/43/44$), so their episode variance is pure shot-sampling noise; Arm~Bs varies with the seed, and its optimization variance is now modeled. Table~\ref{tab:episodes} shows that episode-level std ($0.03$--$0.11$) dwarfs the 3-seed optimization std of Table~\ref{tab:fewshot}; on episode means A $\geq$ C at every budget $n\leq50$, and the small 1-shot arm differences of single-subset runs are inside sampling noise. Seed-averaged paired-episode tests (shared shot draws, episodes 0--9 for Bs contrasts) show C significantly above Bs at $5$- and $10$-shot (95\% CIs $[-0.184,-0.103]$ and $[-0.172,-0.079]$ exclude zero), the 1-shot advantage is not robust to optimization-seed variance (CI $[-0.124,+0.020]$), and the 50-shot Bs-over-C difference of $+0.022$ has CI $[-0.009,+0.053]$ --- a trend, not a significant crossover. Under the looser episode$\times$seed pairing (30 pairs), the 1-shot and 50-shot CIs exclude zero ($[-0.097,-0.006]$ and $[+0.003,+0.041]$); we headline the seed-averaged pairing.

\begin{table}[!htb]
\centering
\footnotesize
\caption{Episode-averaged few-shot sweep on the CC target (macro-F1, mean$\pm$std over episodes; $20$ episodes for A/C, $10$ for Bs, independent stratified shot draws, three optimization seeds $42/43/44$; A/C are deterministic in the seed, so their statistics are unchanged from the single-seed run). ``full'' for Bs is the 3-seed mean of deterministic-budget runs; A/C ``full'' are seed-independent.}
\label{tab:episodes}
\setlength{\tabcolsep}{3pt}
\begin{tabular}{@{}lccccc@{}}
\toprule
Arm & $n{=}1$ & $n{=}5$ & $n{=}10$ & $n{=}50$ & full \\
\midrule
A & $0.511{\scriptsize\pm}.112$ & $0.591{\scriptsize\pm}.080$ & $0.668{\scriptsize\pm}.053$ & $0.707{\scriptsize\pm}.048$ & $0.736$ \\
C & $0.467{\scriptsize\pm}.065$ & $0.576{\scriptsize\pm}.057$ & $0.617{\scriptsize\pm}.063$ & $0.658{\scriptsize\pm}.027$ & $0.705$ \\
Bs & $0.401{\scriptsize\pm}.096$ & $0.443{\scriptsize\pm}.078$ & $0.493{\scriptsize\pm}.105$ & $0.688{\scriptsize\pm}.038$ & $0.792{\scriptsize\pm}.028$ \\
\bottomrule
\end{tabular}
\end{table}

\section{Zero-Shot MLLM Details}
\label{app:mllm}

\textbf{Protocol.} The zero-shot MLLM baseline of Section~\ref{sec:setup} uses \texttt{qwen3-omni-flash} served through the Alibaba DashScope compatible-mode endpoint (snapshot date 2026-07-21), queried once per clip with the raw waveform sent as base64 \texttt{input\_audio} (wav) at temperature $0$, with exponential backoff on 429/5xx and a 1\,s rate limit. The prompt states the five-class need ontology with explicit class definitions and requires the model to answer with exactly one class label; the same ontology classes and macro-F1/UAR/accuracy metrics as B1 are used. Sampling: the CC ($120\to60$) and DAC ($78\to60$) test splits are stratified subsamples by ontology label (seed $42$); BCSD-U ($n=30$) and DI-$\Delta$ ($n=4$) are evaluated in full. Model selection record (reported as-run): \texttt{qwen-audio-turbo-latest} returned HTTP~403 (quota exhausted), \texttt{qwen2.5-omni} returned HTTP~404 (no access), and \texttt{qwen-audio-turbo}/\texttt{qwen2-audio-instruct} are unsupported in compatible mode, so \texttt{qwen3-omni-flash} was used.

\textbf{Results.} Table~\ref{tab:b6} reports per-domain scores. Instruction following was stable: $154/154$ responses parsed to a legal class label (UNPARSED $=0$, refusals $=0$), so the bottleneck is acoustic--semantic representation, not compliance. On DAC the UAR of $0.111$ is below the chance-level UAR of $0.333$: zero-shot predictions collapse toward the majority class, mirroring the low in-domain DAC scores of Table~\ref{tab:b1}. Cost and latency: $154$ clips in total; e.g., BCSD-U $30$ clips $=6{,}930$ prompt tokens $/$ $94$ completion tokens $/$ $132.9$\,s ($\approx4.4$\,s per clip including the rate limit).

\begin{table}[!htb]
\centering
\footnotesize
\caption{Zero-shot \texttt{qwen3-omni-flash} per-domain results (macro-F1 / UAR / accuracy). CC and DAC are stratified 60-clip subsamples of the B1 test splits; DI-$\Delta$ ($n=4$) is directional only. Not comparable cell-for-cell with Table~\ref{tab:b1} (zero-shot vs.\ in-domain supervised probes; subsampled vs.\ full test sets).}
\label{tab:b6}
\setlength{\tabcolsep}{3.5pt}
\begin{tabular}{@{}lccccc@{}}
\toprule
Domain & $n$ & macro-F1 & UAR & acc.\ & unparsed \\
\midrule
BCSD-U & 30 & 0.290 & 0.277 & 0.267 & 0 \\
CC & 60 & 0.217 & 0.197 & 0.250 & 0 \\
DAC & 60 & 0.155 & 0.111 & 0.267 & 0 \\
DI-$\Delta$ & 4 & 0.667 & 0.500 & 0.500 & 0 \\
\bottomrule
\end{tabular}
\end{table}

\textbf{Limitations.} A single MLLM, single prompt, and zero-shot only were evaluated; 5-shot audio exemplars and \texttt{qwen-audio-turbo-latest} were blocked by quota at experiment time. API-model versioning is outside our control; all raw responses are released (\texttt{raw\_predictions.jsonl} per domain) for error analysis.

\section{Encoder and Probe Card}
\label{app:encodercard}

Table~\ref{tab:encodercard} lists the exact frozen-feature configurations. All encoders are read out at the final layer for comparability; layer-wise sensitivity is left to future work. A fifth pre-specified encoder, voc2vec (\texttt{CognitiveReflection/voc2vec}), could not be downloaded (gated repository, HTTP~401 at experiment time); per the pre-specified fallback plan it was replaced by AST as the generic audio-event-pretrained control. The zero-shot MLLM baseline (\texttt{qwen3-omni-flash}; Section~\ref{sec:setup}, Appendix~\ref{app:mllm}) does not pass through this embedding-plus-probe pipeline --- it classifies raw audio directly --- and is therefore not listed in the encoder card.

\begin{table}[!htb]
\centering
\footnotesize
\caption{Encoder card: checkpoints, readout, and dimensions. All audio resampled to 16\,kHz mono, 7\,s, peak-normalized (CLAP internally resampled to 48\,kHz).}
\label{tab:encodercard}
\setlength{\tabcolsep}{2.5pt}
\resizebox{\columnwidth}{!}{%
\begin{tabular}{@{}lllc@{}}
\toprule
Encoder & Checkpoint & Readout & Dim \\
\midrule
WavLM & \texttt{microsoft/wavlm-base-plus} & masked-mean, last hidden state & 768 \\
wav2vec2 & \texttt{facebook/wav2vec2-base} & masked-mean, last hidden state & 768 \\
CLAP & \texttt{laion/clap-htsat-unfused} & audio embedding (\texttt{get\_audio\_features}) & 512 \\
AST & \texttt{MIT/ast-finetuned-}\allowbreak\texttt{audioset-10-10-0.4593} & [CLS] token, last hidden state & 768 \\
handcrafted & librosa (MFCC, prosody, spectral) & mean$+$std over time & 252 \\
\bottomrule
\end{tabular}}
\end{table}

\textbf{Probe and adaptation hyperparameters.}
Linear probes: StandardScaler $+$ logistic regression, class-weight balanced, $C\in\{0.01,0.1,1,10\}$ selected on the source-domain validation split by macro-F1 (saga solver for the in-domain B1 runs; lbfgs for all transfer runs), max $5{,}000$ iterations. After an external audit found the selection loop was not passing $C$ to the estimator (all runs silently at $C{=}1$), the shared trainer was fixed, unit tests were added (grid candidates change the fitted estimator; selection is non-degenerate across seeds), and every probe-dependent result was re-run from cached embeddings; the values reported here are from the corrected runs, in which non-default $C$ values are genuinely selected (e.g., $C{=}10$ for mapped joint training; $C{=}0.01$ for many transfer cells). XGBoost baseline: histogram-based gradient boosting with balanced sample weights, default depth/learning rate. Arm B (full fine-tuning): lr $10^{-4}$, bf16, at most 5 epochs, early stopping on validation. Arm Bs (stabilized): convolutional feature encoder frozen, transformer stack at lr $10^{-5}$. Arm C (DAPT): wav2vec2 masked-prediction continuation on $4{,}537$ unlabeled cry clips (corpus train splits $1{,}037$ $+$ CryCeleb $3{,}500$; inductive protocol, Appendix~\ref{app:protocol}), one epoch, 283 steps. Tuning budget was held identical across encoders and arms; no per-encoder hyperparameter search was performed.

\section{Quality--Transfer Details}
\label{app:quality}

\textbf{Per-pair correlations.}
Of the $30$ directed encoder$\times$pair correlations, $21$ are negative. The nine positive pairs comprise all five CC$\to$DAC pairs ($+0.05$ to $+0.10$), all three handcrafted CC/BCSD-U pairs, and AST CC$\to$BCSD-U ($+0.042$). The strongest negative values occur with DAC as \emph{source} (CLAP DAC$\to$BCSD-U $r=-0.334$; wav2vec2 DAC$\to$BCSD-U $r=-0.286$; WavLM DAC$\to$BCSD-U $r=-0.264$; all $p<10^{-3}$). Per-encoder ranges span $[-0.334,+0.050]$ (CLAP). Domain-level descriptives (no test, $n=3$ domains): the highest-SNR domain is the hardest transfer target (BCSD-U $102.9$\,dB $\to$ mean incoming cross-domain F1 $0.496$; CC $37.2 \to 0.531$; DAC $32.3 \to 0.524$; CC and DAC nearly tied).

\textbf{Estimator validity.}
Our SNR estimator is not comparable across sampling rates: on the clean 44.1\,kHz BCSD-U recordings it saturates ($102.9$\,dB is a numerical artifact, not a physical SNR), so pooled cross-domain correlations conflate estimator behavior, sampling rate, and domain identity. Within-domain recomputation (Table~\ref{tab:withinsnr}), where the estimator is at least self-consistent, shows at most weak correlations: none significant on DAC; two of five positive and significant on CC. Fisher's combined-$p$ over per-pair tests is also anti-conservative here because per-clip correctness is clustered by domain pair; we therefore report it only as a descriptive summary.

\begin{table}[!htb]
\centering
\small
\caption{Within-domain point-biserial correlation between clip-level SNR and prediction correctness (pooled over seeds). No correction across the 10 tests; $^*$ $p<0.05$ uncorrected.}
\label{tab:withinsnr}
\setlength{\tabcolsep}{4pt}
\begin{tabular}{@{}lcc@{}}
\toprule
Encoder & DAC & CC \\
\midrule
WavLM & $-0.075$ ($p=.25$) & $+0.111$ ($p=.036$)$^*$ \\
wav2vec2 & $+0.041$ ($p=.53$) & $-0.014$ ($p=.79$) \\
CLAP & $+0.051$ ($p=.43$) & $+0.127$ ($p=.016$)$^*$ \\
AST & $-0.042$ ($p=.52$) & $+0.077$ ($p=.14$) \\
handcrafted & $+0.043$ ($p=.51$) & $-0.032$ ($p=.55$) \\
\bottomrule
\end{tabular}
\end{table}

\textbf{BCSD-U ceiling confidence bounds.}
The BCSD-U test split has $n=30$; after the corrected-$C$ rerun two encoders tie at macro-F1 $1.000$ (CLAP, AST), and WavLM no longer sits at a perfect score (pooled accuracy $0.967$). Table~\ref{tab:bcsduci} gives Clopper--Pearson 95\% lower bounds on accuracy: even a perfect $30/30$ cell is consistent with true accuracy as low as $0.884$ (single seed) / $0.960$ (pooled over seeds), so BCSD-U diagonal cells should be read as ``near-ceiling, resolution-limited'' rather than as exact zeros of error.

\begin{table}[!htb]
\centering
\footnotesize
\caption{BCSD-U diagonal cells: accuracy and Clopper--Pearson 95\% lower bounds (single seed $n=30$; pooled $n=90$). The pooled macro-F1 is computed once on the pooled 90-clip prediction set (3 seeds combined); it therefore differs slightly from the seed-averaged macro-F1 of Table~\ref{tab:b1} (mean of per-seed values, e.g., wav2vec2 $0.833$ pooled vs.\ $0.883$ seed-averaged; handcrafted $0.944$ vs.\ $0.961$). Both conventions are reported as computed.}
\label{tab:bcsduci}
\setlength{\tabcolsep}{2.5pt}
\resizebox{\columnwidth}{!}{%
\begin{tabular}{@{}lcccc@{}}
\toprule
Encoder & pooled macro-F1 & acc.\ (pooled) & LB (pooled) & LB (1 seed) \\
\midrule
WavLM & 0.944 & 87/90 & 0.906 & 0.828 \\
wav2vec2 & 0.833 & 81/90 & 0.819 & 0.735 \\
CLAP & 1.000 & 90/90 & 0.960 & 0.884 \\
AST & 1.000 & 90/90 & 0.960 & 0.884 \\
handcrafted & 0.944 & 87/90 & 0.906 & 0.828 \\
\bottomrule
\end{tabular}}
\end{table}

\begin{figure}[t]
\centering
\includegraphics[width=0.92\columnwidth]{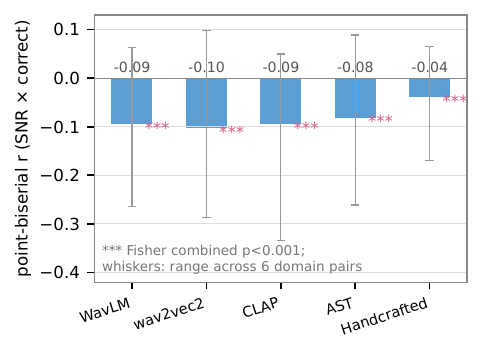}
\caption{\textbf{Clip-level SNR negatively correlates with cross-domain correctness — a confounded signal.} Point-biserial correlation $r$ between estimated SNR and cross-domain prediction correctness, pooled over the six ordered domain pairs per encoder (whiskers: min--max across pairs; stars: Fisher combined $p$). All encoders show a significant negative mean correlation; however, the estimator saturates on 44.1\,kHz audio (BCSD-U 102.9\,dB is an artifact) and SNR is almost perfectly confounded with domain identity, so this effect cannot be interpreted as a causal quality effect (C5 refuted as stated).}
\label{fig:quality}
\end{figure}

\section{Within-Corpus Train--Test Near-Duplicate Audit}
\label{app:withinaudit}

This section reports the per-corpus train--test leakage audit summarized in Section~\ref{sec:within} (script \texttt{audit\_within\_corpus.py}; all outputs under \texttt{results/audit\_within\_corpus/}).

\textbf{Group semantics.}
DAC: \texttt{group\_id} is the \texttt{uuid} filename prefix (uploader/device ID; prefix match rate $100\%$) --- same device with different epoch timestamps counts as different recordings, so recording identity is not captured. BCSD-U: the group is the clip ID itself ($207$ clips / $207$ groups). CC: the file stem with the \texttt{Uncom\_Rev\_} wrapper removed; $83$ size-2 groups, of which $78$ are explicit \texttt{<X>.wav} $+$ \texttt{Uncom\_Rev\_<X>.wav} re-export pairs (median group size $1$). No group crosses splits in any corpus, but group $\neq$ infant/recording identity, so cross-group same-source pairs can and do span the train--test boundary.

\textbf{Hits by method.}
Exact sha1: $0$ hits in all three corpora. Perceptual fingerprint (64-bit log-mel simhash, hamming $\leq1$, threshold calibrated on the min-distance distribution): DAC $8$, BCSD-U $12$, CC $42$ pairs --- all cross-group. Embedding nearest neighbor (wav2vec2 cosine, top-1 per test clip): pairs above $0.995$ are label-consistent (BCSD-U $2$ pairs; CC $7$ pairs; DAC $0$). Time-shift matching (normalized waveform cross-correlation, threshold $0.80$, computed on fingerprint/embedding candidates): DAC $0$; BCSD-U $5$ pairs, all time-shifted ($\pm1$--$2$\,s) slices of the same long recording \texttt{laugh\_1.m4a} spanning train/test; CC $13$ pairs, all offset $0$ with xcorr $\approx1.000$ --- waveform-identical re-exports whose sha1 differs, invisible to exact hashing. We state this CC finding explicitly as a residual leakage channel that the released manifest's byte-level audit cannot catch.

\textbf{Removal and recomputation.}
The suspicious test clips (union of all four methods: $8/13/44$ for DAC/BCSD-U/CC, $65$ total) were removed and the full $5$-encoder matrix recomputed under the unified protocol on the cleaned test splits. Table~\ref{tab:withinnudtr} reports the diagonal and NTR changes. The diagonal moves by $\Delta\in[-0.072,+0.015]$ (largest drop AST/DAC $0.521\to0.449$ on an 8-clip removal; three wav2vec2 cells move up), so the diagonal is not systematically inflated; cleaned NTR remains significantly positive for all five encoders. Known limits: time-shift matching is computed only on fingerprint/embedding candidates (full pairwise waveform cross-correlation is too expensive), the fingerprint is a self-implemented simhash proxy rather than chromaprint, and DAC's 8 hits rest on a single method (lower confidence).

\begin{table}[!htb]
\centering
\footnotesize
\caption{Within-corpus audit: diagonal macro-F1 and NTR before/after removing the $65$ suspicious test clips (unified protocol, corrected $C$ selector). NTR stays positive for every encoder; no diagonal drop exceeds $0.072$.}
\label{tab:withinnudtr}
\setlength{\tabcolsep}{2.5pt}
\resizebox{\columnwidth}{!}{%
\begin{tabular}{@{}lccccc@{}}
\toprule
Encoder & DAC diag & BCSD-U diag & CC diag & NTR orig & NTR clean \\
\midrule
WavLM & $0.422\to0.417$ & $0.944\to0.924$ & $0.730\to0.732$ & 0.194 & 0.198 \\
wav2vec2 & $0.571\to0.577$ & $0.833\to0.837$ & $0.736\to0.751$ & 0.200 & 0.200 \\
CLAP & $0.508\to0.506$ & $1.000\to1.000$ & $0.829\to0.783$ & 0.330 & 0.337 \\
AST & $0.521\to0.449$ & $1.000\to1.000$ & $0.798\to0.769$ & 0.345 & 0.323 \\
handcrafted & $0.435\to0.431$ & $0.944\to0.924$ & $0.816\to0.763$ & 0.271 & 0.207 \\
\bottomrule
\end{tabular}}
\end{table}

\section{Reproducibility}
\label{app:repro}

We release: ontology mapping rules as executable code; the deduplication manifest (all $1{,}247$ removed pairs); group-stratified split files; the audit pipeline (feature extraction, probes, bootstrap, within-corpus near-duplicate audit); the near-duplicate pair list, removal-robustness matrices, and the $65$-clip within-corpus exclusion list; per-run estimator parameters (including the validation-selected $C$ of every run) and software versions for every number reported; and the corrected-trainer unit tests. The released manifest \emph{retains} the $312$ clips involved in suspicious near-duplicate pairs (they are label-consistent and their removal does not change any conclusion); the pair list ships as an exclusion file so that the removal-robustness variant of every table can be reproduced exactly. Reproducibility is layered: the audit tables, conflict lists, and manifests can be reverified without raw audio; feature extraction and retraining require obtaining the corpora. Logistic-regression rows are deterministic given features and splits, so their 3-seed std is $\approx0$ by construction; variance estimates are meaningful only for the XGBoost and fine-tuning arms. Licensing, itemized: DAC (ODbL-1.0) and DI-$\Delta$ (Apache-2.0) permit redistribution but we still release no audio; BCSD-U and CC ship without explicit licenses and their original distribution channels could not be fully verified, so both are used \emph{for evaluation only} --- for these we release only sha1 manifests, split files, and label lists so that lawful holders of the original files can verify byte-for-byte; CryCeleb was used as unlabeled DAPT audio under the license terms as stated by its distributors (CC BY-NC-ND 4.0, research use only), and we release no CryCeleb audio and no derivative model weights trained on it (the DAPT checkpoint is not distributed). The zero-shot MLLM raw responses (\texttt{raw\_predictions.jsonl} per domain) are released for error analysis.